\documentclass{article}
\PassOptionsToPackage{numbers, compress}{natbib}
 \usepackage[preprint]{neurips_2025}

\usepackage[utf8]{inputenc} % allow utf-8 input
\usepackage[T1]{fontenc}    % use 8-bit T1 fonts
\usepackage{hyperref}       % hyperlinks
\usepackage{url}            % simple URL typesetting
\usepackage{booktabs}       % professional-quality tables
\usepackage{tabularx}
\usepackage{multirow} 
\usepackage{amsmath} 
\usepackage{rotating}
\usepackage{amsfonts}       % blackboard math symbols
\usepackage{nicefrac}       % compact symbols for 1/2, etc.
\usepackage{microtype}      % microtypography
\usepackage{xcolor}         % colors
\usepackage[capitalize,noabbrev]{cleveref}
\usepackage{graphicx}
\usepackage{threeparttable} 
\usepackage{pifont}
 \usepackage{caption}
\usepackage{enumitem}
\usepackage{subcaption}

\hypersetup{
  colorlinks=true,      % 开启彩色链接（不带框）
  citecolor=blue,       % 文献引用的颜色，如 [3,4]
  linkcolor=red,     % 内部链接（section, figure, table等）的颜色
  urlcolor=blue         % URL 链接颜色
}

\newcommand{\update}[1]{{\textcolor{black}{#1}}}

\newcommand\graycross{\textcolor[rgb]{ .502,  .502,  .502}{\ding{55}}}
\newcommand{\cmark}{\ding{51}}  % ✓
\newcommand{\xmark}{\ding{55}}  % ✗

\title{Teaching Time Series to See and Speak: Forecasting with Aligned Visual and Textual Perspectives}
\author{
Sixun Dong\textsuperscript{1} \hspace{1.5em}
Wei Fan\textsuperscript{2} \hspace{1.5em}
Teresa Wu\textsuperscript{1} \hspace{1.5em}
Yanjie Fu\textsuperscript{1}\thanks{Corresponding author.} \\
\textsuperscript{1}Arizona State University \quad 
\textsuperscript{2}University of Oxford \\
\texttt{\{sixun.dong, teresa.wu, yanjie.fu\}@asu.edu}, \texttt{wei.fan@ox.ac.uk}
}

\begin{document}

\maketitle

\begin{abstract}

Time series forecasting traditionally relies on unimodal numerical inputs, which often struggle to capture high-level semantic patterns due to their dense and unstructured nature. While recent approaches have explored representing time series as text using large language models (LLMs), these methods remain limited by the discrete nature of token sequences and lack the perceptual intuition humans typically apply, such as interpreting visual patterns.
In this paper, we propose a multimodal contrastive learning framework that transforms raw time series into structured visual and textual perspectives. Rather than using natural language or real-world images, we construct both modalities directly from numerical sequences. We then align these views in a shared semantic space via contrastive learning, enabling the model to capture richer and more complementary representations.
Furthermore, we introduce a variate selection module that leverages the aligned representations to identify the most informative variables for multivariate forecasting. Extensive experiments on fifteen short-term and six long-term forecasting benchmarks demonstrate that our approach consistently outperforms strong unimodal and cross-modal baselines, highlighting the effectiveness of multimodal alignment in enhancing time series forecasting. Code is available at: \url{https://github.com/Ironieser/TimesCLIP}.

\iffalse
% Wei
Time series forecasting aims to predict future data points by analyzing historical data. While numerous methods treat time series as unimodal data, many overlook how humans intuitively understand time series through visual signals rather than solely numerical data points. The miss of critical aspects hidden in visual representations, as well as the ignorance of multimodal modeling, could largely hinder the performance of time series forecasting.
Drawing inspiration from multimodal learning in computer vision and natural language processing, we introduce a novel multimodal contrastive learning framework for time series forecasting that converts time series data into visual and language representations. This approach enhances the model's forecasting capabilities by aligning time series with multimodal space that encapsulate varied dimensional information. Moreover, we design a novelty variate selection module to leverage multimodal feature, which could figure out the most significant variate from multiple variables of time series. Extensive experiments demonstrate that our method achieves state-of-the-art performance on one short-term and four long-term time series forecasting datasets.
\fi
\end{abstract}

%\vspace{-2.5em}
\section{Introduction}
Time series data, consisting of records captured over time, are prevalent in our lives and various application domains~\cite{li2024deep}. 
%such as electricity~\cite{}, weather~\cite{}, stock markets~\cite{}, biomedicine~\cite{}, and national finance~\cite{}. 
%In healthcare, for instance, physicians utilize electrocardiograms (ECG) and electroencephalograms (EEG) to analyze heart and brain activity for diagnosing diseases~\cite{}. 
%Similarly, financial analysts examine trends in stock market data to develop high-yield investment strategies~\cite{}. 
As one of the essential tasks in time series data, time series forecasting, involves predicting future data points by analyzing historical data. 
Given its widespread demand in real-world scenarios, time series forecasting has attracted considerable research interest from the artificial intelligence and deep learning community, particularly in fields such as meteorology \cite{zhang2024integration,wu2023interpretable,zhang2023skilful}, energy \cite{weron2014electricity,li2024deep,runge2021review}, healthcare \cite{kaushik2020ai,bui2018time}, and financial investment \cite{cheng2022financial,lu2021cnn}.

\begin{figure*}[ht] 
\centering
\begin{minipage}{0.5\textwidth} 
    \centering
    \includegraphics[width=\textwidth]{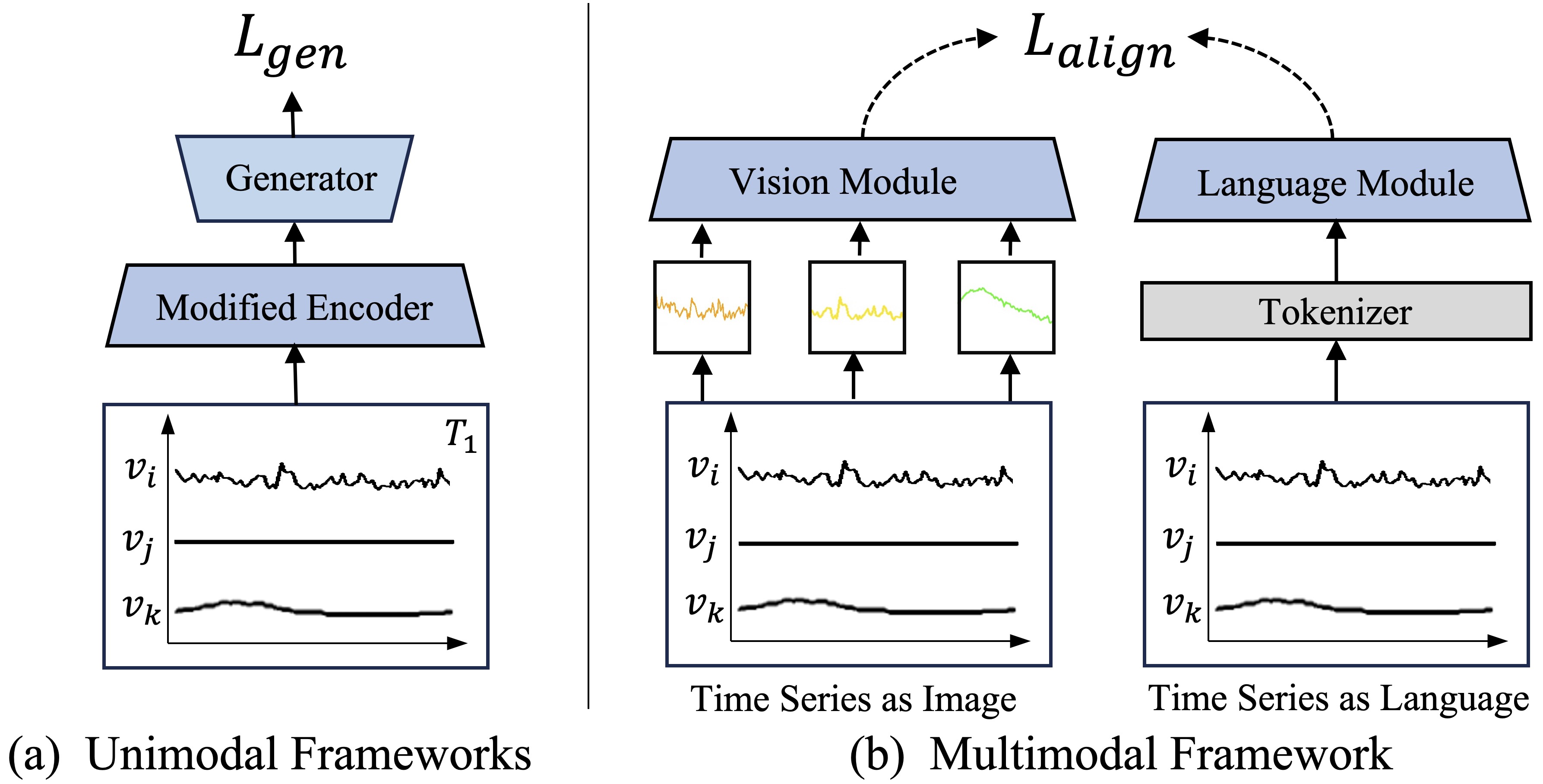}
    \vspace{-0.5em}
    % \subcaption{Unimodal vs. Multimodal Frameworks.}
\end{minipage}
\hfill
\begin{minipage}{0.48\textwidth}
    \centering
    \scriptsize
    \resizebox{\textwidth}{!}{
    % \begin{tabular}{l|c|ccc|c|c}
    % \toprule
    % \textbf{Method} & \textbf{Raw Input} & \textbf{LM} & \textbf{Train Embed.} & \textbf{LM Usage} & \textbf{Vision} & \textbf{MM Feature} \\
    % \midrule
    % Time-LLM        & Num. (prompted)    & LLaMA      & \xmark & LM-based TSF.  & \xmark & \xmark \\
    % Chronos (T5)    & Num. (tokenized)   & T5         & \cmark & LM-based TSF.  & \xmark & \xmark \\
    % PatchTST        & Numerical          & \xmark     & \cmark & \xmark                & \xmark & \xmark \\
    % iTransformer    & Numerical          & \xmark     & \cmark & \xmark                & \xmark & \xmark \\
    % \textbf{TimesCLIP} & \textbf{Numerical} & CLIP-Text & \cmark & Feature-only         & \cmark & \cmark \\
    % \bottomrule
    % \end{tabular}
    \begin{tabular}{l@{\hskip 4pt}c@{\hskip 4pt}c@{\hskip 4pt}c@{\hskip 4pt}c@{\hskip 4pt}c@{\hskip 4pt}c}
        \toprule
        \textbf{Method} & \textbf{Input} & \textbf{LM} & \textbf{Train Emb.} & \textbf{LM Use} & \textbf{Vision} & \textbf{MM Feat.} \\
        \midrule
        Time-LLM        & Prompted     & LLaMA      & \xmark &LM-based TSF. & \xmark & \xmark \\
        Chronos (T5)    & Tokenized    & T5         & \cmark &LM-based TSF. & \xmark & \xmark \\
        PatchTST        & Numerical    & \xmark     & \cmark & \xmark        & \xmark & \xmark \\
        iTransformer    & Numerical    & \xmark     & \cmark & \xmark        & \xmark & \xmark \\
        \textbf{TimesCLIP} & \textbf{Numerical} & CLIP-Text & \cmark &Extract Feat.& \cmark & \cmark \\
        \bottomrule
        \end{tabular}

    }
    \vspace{2em}
    \text{(c) Comparison of input modalities and modeling strategies.}
    % \vspace{0.5em}
    % \subcaption{Comparison of modeling paradigms and input strategies.}
\end{minipage}
\vspace{-1em}
\caption{
Comparison of frameworks and modeling strategies. Overview of modeling paradigms for Time Series Forecasting(TSF). The left figure contrasts unimodal and multimodal frameworks, while the right table summarizes representative methods across core modeling dimensions. TimesCLIP uniquely constructs and aligns visual and textual features directly from numerical time series, enabling multimodal learning without requiring external modalities.
}
% \label{fig:teaser_plus_table}
\label{fig:teaser}
\vspace{-2em}
\end{figure*}

Classic deep learning-based approaches to time series forecasting usually rely on designing unique architectures as different end-to-end forecasting models.  
Researchers have explored various architectures, including Transformer-based \cite{patchtst,liu2023itransformer}, CNN-based \cite{wang2023micn}, MLP-based \cite{N-BEATS,zhang2022less}, and RNN-based models \cite{lin2023segrnn,liu2024autotimes}, to accomplish forecasting. 
However, these methods mostly treats time series data purely as numerical sequences; in other words, they can be depicted as \textbf{\textit{unimodal}} frameworks, as shown in \cref{fig:teaser}(a).
Unfortunately, unimodal frameworks are limited in modeling complex patterns, contextual semantics, and long-term dependencies, and, thus, often result into fragile models.
%Since raw \textit{unimodal} time series contain unprocessed, dense, and noisy data points, it is challenging for models to extract critical information, particularly the temporal dependencies necessary for forecasting. 
%Although some models, like TimesNet \cite{Timesnet}, transform numerical data into the frequency domain, they still focus primarily on a single modality and do not utilize human intuitive knowledge, as demonstrated by doctors diagnosing conditions using ECG.

With the success of Large Language Models (LLMs), researchers in time series have brought up the new perspective of \textbf{\textit{cross modality}}. 
There are studies that treat time series as a ``foreign language'' that can be decoded using the language representation strategies of large language models through fine-tuning~\cite{sun2023test,liu2024lstprompt,liu2024autotimes} or reprogramming dedicated time series encoders \cite{jin2023time}. 
Despite these advances, representing time series as a language is still less intuitive, as humans typically interpret time series data \textit{\textbf{visually}} rather than \textit{\textbf{textually}}~\cite{tan2024language}. For example, doctors usually adopt the visual information in ECG to diagnose heart conditions~\cite{ribeiro2020automatic}; financial professionals would conduct analysis based on the stock market charts instead of pure numbers~\cite{bisiotis2022control}. 
Recent studies have suggested that directly applying LLM techniques to time series forecasting may not always yield effective results \cite{tan2024language,jin2024position}. 
% While some works have explored converting time series to images \cite{li2020forecasting,semenoglou2023image,barra2020deep,li2024time,li2024lite}, they have generally underperformed in time series forecasting \cite{semenoglou2023image,li2020forecasting} or have relied on Fourier-transformed spectrograms \cite{barra2020deep,wang2015imaging} rather than visual representations.   Additionally, \cite{liu2024focal,cai2024orthogonality} attempt to consider different sensors as multimodal data, however, these sensors record same type data, human motion, not a common definition.
Recent cross-modality studies~\cite{li2020forecasting,semenoglou2023image,barra2020deep,li2024time,li2024lite} have converted time series data into images for diverse tasks, but are not performant in time series forecasting~\cite{semenoglou2023image,li2020forecasting}. 
More recent studies have utilized Fourier-transformed spectrograms \cite{barra2020deep,wang2015imaging}, which, however, don't provide visually interpretative time series representations. 

Besides, efforts have been made to extract \textit{\textbf{multimodal}} knowledge for time series related domain-specific applications. 
The studies in~\cite{liu2024focal,cai2024orthogonality} attempted to classify different sensors; however, the sensors capture the same type of data-human motion, which does not conform to the true multimodality. 
In healthcare, the integration of medical time series and disease description texts to construct multimodal frameworks has been explored \cite{ribeiro2020automatic}.
However, these applied studies benefit domain-specific data with annotated multimodal information, but are limited in generalized time series analysis and forecasting.

In response to the above challenges, we introduce \textit{TimesCLIP}, a novel Multimodal Contrastive Learning approach for time series modeling and forecasting. 
Inspired by the effective modeling of CLIP~\cite{CLIP} that bridges the gap between text and visual understanding, \textit{TimesCLIP} introduces a novel idea to jointly learn both numeric time series and visual representations.
Specifically, it involves converting time series data into images and employing a multimodal contrastive learning framework. \cref{fig:teaser} (b) shows that, in the vision branch, original numerical variates are transformed into distinct figures using different colors. In the language branch, similar to LLM-based approaches for time series, we treat numerical time series data as a "foreign" language. However, instead of employing a complex projection module to align time series with "real" language, we employ a simpler learning tokenizer to align time series with the feature space of pretrained language model. 
We further introduce a modified contrastive loss specifically designed for multimodal time series, aiming to align the vision representation with the language representation within a multimodal space. 
Subsequently, we introduce an innovative variate selection module. 
We designate the aligned classification feature as ``Query'' and use a cross-attention layer to identify the most correlative features for downstream tasks in the multivariate feature sequence. This module enables the model to effectively utilize multimodal features and classification information to identify the most significant one among multiple variables for downstream tasks.
In brief,
% Consequently, to evaluate the effectiveness of our proposed model, TimesCLIP, we conduct extensive experiments on two time series tasks: long-term forecasting and short-term forecasting. The results show that our approach outperforms other baselines and verify our proposed method effectiveness.
we summarize our contribution in threefold:
% \begin{itemize}
%     \setlength{\itemsep}{0pt}  % 适当增加间距
%     \item We introduce a novel concept of converting numerical multivariate time series to vision and language representation. We also propose a novelty multimodal contrastive learning framework for time series to align time series with vision-language multimodal space.
%     \item We propose a variate selection module to leverage the aligned multimodal feature to identify the most corrective variable among multivariate of time series for time series forecasting.
%     \item Extensive experiments show our proposed model, TimesCLIP, achieve strong performance on short-term forecasting and long-term forecasting.
% \end{itemize}

% \begin{itemize}
%     \item We present a novel concept of converting numerical multivariate time series into vision and language representations. Additionally, we propose an innovative multimodal contrastive learning framework to align time series data with vision-language multimodal spaces.
%     \item We introduce a variate selection module designed to leverage aligned multimodal features, enabling the identification of the most relevant variable among the multivariate time series for forecasting purposes.
%     \item Extensive experiments demonstrate that our proposed model, TimesCLIP, achieves strong performance in both short-term and long-term forecasting tasks.
% \end{itemize}
\vspace{-1em}
\begin{itemize}[leftmargin=*]
    \item We present a novel concept for converting numerical multivariate time series into vision and language representations. We propose an innovative multimodal contrastive learning framework to align time series data with a vision-language multimodal space.
\vspace{-0.5em}
    \item We introduce a variate selection module designed to leverage aligned multimodal features, enabling the identification of the most relevant variable among the multivariate time series for time series forecasting.
    \vspace{-0.5em}
    \item Extensive experiments demonstrate that our proposed model, TimesCLIP, achieves strong performance on both short-term and long-term forecasting tasks.
\end{itemize}

% \begin{compactitem}
%     \item We propose a novelty multimodal contrastive learning framework for time series forecasting, which can align time series with vision-language multimodal space.
    
%     \item We introduce a variate selection module to leverage the aligned multimodal feature to figure out the signification variable for time series forecasting for multivariate time series data.
%     \item Extensive experiments show our proposed model, TimesCLIP, achieve strong performance on short-term forecasting and long-term forecasting.
% \end{compactitem}

% \section{Introduction}

% With contrastive learning\cite{chen2020simple} rising up recently, some researchs are trying 

\vspace{-0.4cm}
\section{Related Work}
\vspace{-0.2cm}
\noindent\textbf{Multimodal Learning for Time Series.}
% \vspace{-0.1cm}
Contrastive learning has significantly advanced the processing and understanding of multimodal data, especially CLIP~\cite{CLIP}, which illustrates the generalization ability of multimodal contrastive learning.  \cite{yang2017deep} built deep multimodal representation learning from temporal data, which utilizes different modality information, such as audio, image, and language.  \cite{shi2024dust} introduced the Dual Swin Transformer (DuST), a multimodal classification framework, that integrates video with synchronous time series data for driving risk assessment. \cite{li2024frozen} explored a frozen language model for zero-shot learning on ECG data. \cite{ebrahimi2023lanistr} integrated structured and unstructured data as different modalities. However, these works require the original dataset to contain more than one modality or do not require aligning different modalities. The benchmarks for time series forecasting \cite{Timesnet,liu2023itransformer}, only contain recorded numerical data, cannot work with these methods. We summarize representative forecasting paradigms and their use of language models, input modalities, and multimodal strategies in Figure~\ref{fig:teaser}(c). \cite{liu2024focal,cai2024orthogonality} claim that they attempt to utilize multimodal learning for human motion classification, a type of time series data. However, they remain constrained by the use of various sensors placed at different positions on the human body for action recognition. They do not adhere to the commonly accepted definition of multimodality in the AI field, which traditionally involves integrating distinct modalities such as vision, audio, language, and depth, rather than using different sensors to record the same modality information. While recent work has explored applying vision-language models (VLMs) to time series data~\cite{timevlm}, we instead propose a contrastive alignment method specifically designed for numerical time series.

In addition, we include more related work on time series forecasting with language models, vision-language contrastive learning, and time series foundation models in \cref{sec:app_related}. 

% Several datasets have emerged as benchmarks to advance research by providing standardized frameworks for testing and validation. \textit{Time-MMD}\cite{liu2024time} is a multi-domain multimodal dataset designed for time series analysis. \textit{Bjtt}\cite{zhang2024bjtt} focuses on traffic prediction \cite{zhang2024bjtt}, and \textit{PTB-XL}\cite{wagner2020ptb} serves as a standard for cardiac signal analysis \cite{wagner2020ptb}. However, these benchmarks are not 

% \subsection{Time Series Representation}

% Extensive studies have explored time series representation and pre-training strategies\cite{patchtst,}

% \vspace{-0.2cm}
\section{Method}

In this section, we first present our proposed model in \cref{sec:overview} for time series forecasting. We explain the modified vision representation module and language module in \cref{sec:vision-language} and contrastive learning multimodal alignment for time series in \cref{sec:loss_cont}, followed by the designed variate selection module in \cref{sec:var}. The aligned language representation of time series data is passed into the generator to forecast future time series data in \cref{sec:generator}.

% \vspace{-0.1cm}
\begin{figure*}[t]
% \vskip 0.2in
\begin{center}
% \vspace{-0.5em}
\centerline{\includegraphics[width=0.75 \textwidth]{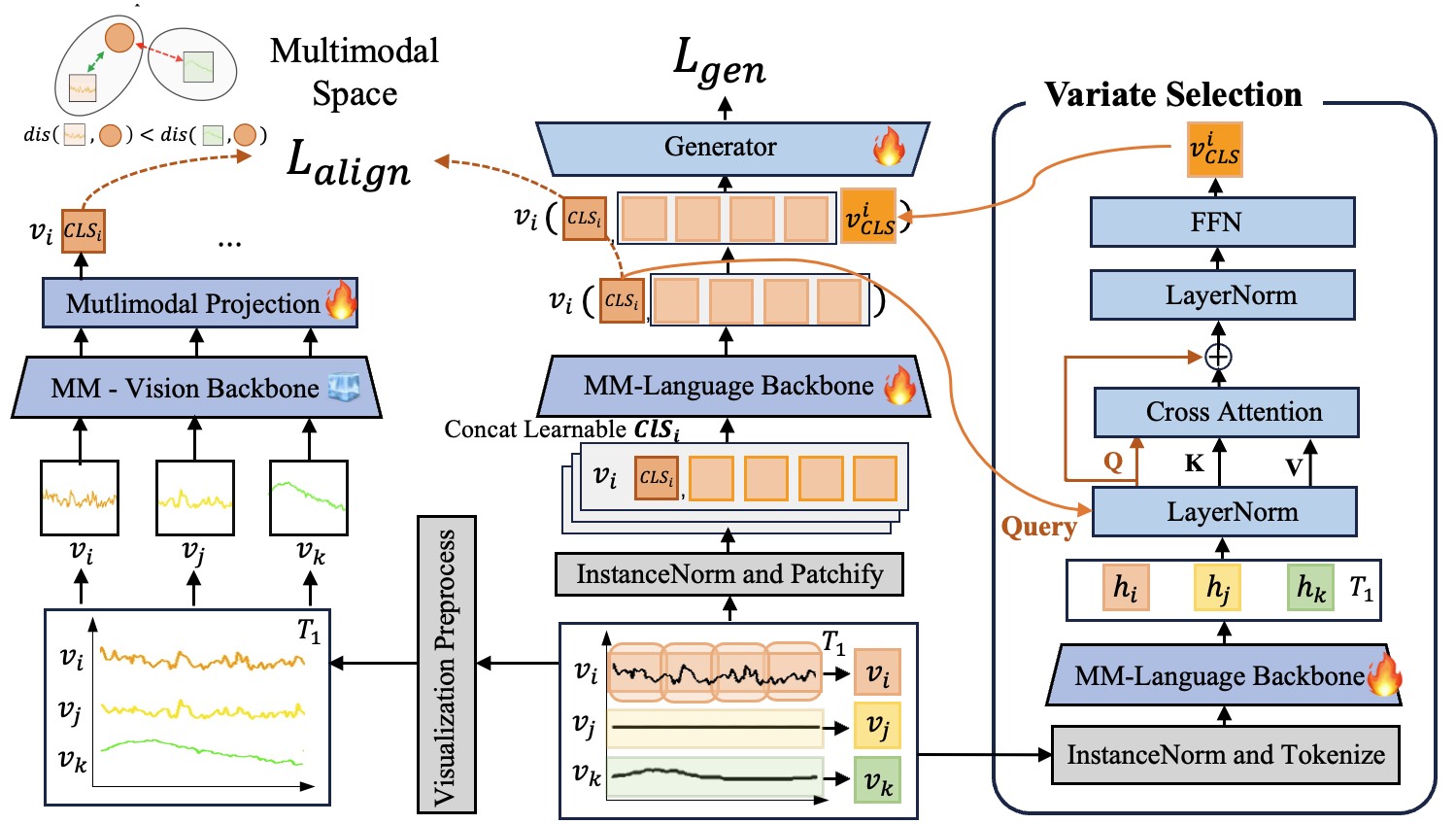}}
 
% \vspace{-1em}
\caption{
% Overview of the proposed TimesCLIP. This framework includes two primary modules: the Vision Module and the Language Module. The Vision Module begins by converting time series data into colorized figures, extracting visual representations using a pre-trained vision backbone, and applying a learnable projection layer for alignment with the language space. The Language Module processes multivariate time series by patchifying them into patches, appending an additional $class$ token to the sequence, and then feeds this sequence into a pre-trained language encoder to obtain the language representation of time series. Subsequently, a contrastive learning loss is employed to align the vision representation of the time series with its language $class$ token. A variate selection module then identifies the most correlative variates for the downstream task in the multivariate time series. Finally, an MLP-based generator generates future time series data.
 Overview of the TimesCLIP framework. TimesCLIP converts numerical multivariate time series into colorized figures and patchified sequences, which are then processed by pretrained vision and language backbones. A learnable $\texttt{CLS}$ token is appended on the language side to serve as a unified representation for contrastive alignment. A multimodal InfoNCE loss $\mathcal{L}_{\text{align}}$ aligns vision and language features into a shared space. The aligned $\texttt{CLS}$ token is then used to identify task-relevant variates via cross-attention, and passed to a generator for forecasting.
}
\vspace{-2.5 em}
\label{fig:framework}
\end{center}
\end{figure*}
% \begin{figure}[ht!]
% % \vskip 0.2in
% \begin{center}
% \centerline{\includegraphics[width=0.5\textwidth]{fig/framework1.jpg}}
% \caption{Proposed Framework}
% \label{fig:framework}
% \end{center}
% \end{figure}

\subsection{Overview}
\label{sec:overview}
\cref{fig:teaser} shows our general-purpose architecture of multi-modal time series contrastive learning includes two modules: (i) vision module and (ii) language module.
To implement the general purpose architecture, \cref{fig:framework} shows our proposed model TimesCLIP. Compared with the proposed general multimodal architecture in \cref{fig:teaser}, TimesCLIP consists of three parts: (i) a multimodal vision module, (ii) a multimodal language module, and (iii) a variate selection module. 

The left part of \cref{fig:framework} shows the vision module of  TimesCLIP contains three blocks: the Visualization Preprocess for time series, a pretrained and frozen multimodal vision backbone to extract time series figure features, and a trainable multimodal projection layer.
First, the Visualization Preprocess converts the original time series signals into figures, by using a designed normalization step to normalize the original numerical values and visualize them in the figures with different colors. 
By converting the original time series into images, we can utilize existing image feature encoders to extract time series features from a visual perspective. Additionally, following \cite{vit,coca,CLIP}, we feed vision features to a learnable multimodal  projection to transform and represent time series vision features into a multimodal time series feature space. Overall, this vision branch plays the most important role in our proposed multimodal framework, which successfully converts the numerical data of the time series into a visual representation space that is to be aligned with language representation space. 

The right part of \cref{fig:framework} shows that we patchify each variate of time series data and leverage an online learnable time series tokenizer to project the patchified time series signal into language tokens. After concatenating classification class tokens for each variate, we pass them into the pretrained language encoder to obtain the language representation of time series data. Additionally, the introduced class tokens are utilized for multimodal alignment. More explanation about the modules above is in \cref{sec:vision-language}. 
Specifically, we introduce a special variate selection module, which utilizes classification information from cross-modal space to select the most significant correlative variate feature from multi-variate time series data. 
Finally, we feed the aligned language representation into the generator to produce the final forecast.

% \vspace{-0.1cm}
\subsection{Vision-Language Module for Time Series}
\label{sec:vision-language}
In time series forecasting, given a length-$T$ historical multivariate time series observation $X^{1:T}=\{ v_1^{1:T},\dots,v_N^{1:T}\}\in\mathbb{R}^{T\times N}$, $v_i^{1:T} =\{v_i^1,\dots,v_i^T\}, i\in[1,2,\dots,N]$. $N$ represents different recorded variates, and $T$ is the known historical time step. We predict the future $S$ time steps of the time series $Y=X^{1+S:T+S} = \{v_1^{1+S:T+S},\dots,v_N^{1+S:T+S}\}$. 

\cref{fig:framework} shows that, with different preprocessing procedures for time series data, its vision representation and language representation are produced by the vision module and language module, respectively. 

\noindent \textbf{Vision module}.
Since original time series data is numerical signals with multivariate,  differences in numerical ranges exist. To visualize these data with reasonable value ranges in the figure, we normalize every variable in a fixed window size $L_{window}$ instead of normalizing in all time series data or mini-batch data. In this way, we reduce the impact of outlier maxima or minima compared to normalizing the values in the minibatch or all data. Next, we plot each time series variable with a specific color to help discriminate variable types to align in the multimodal feature space with the language representation of time series. This process can be denoted as \cref{eq:vis}. 
\begin{equation}
    I_i = \text{Visualize}(v_i^{1:T}), i\in[1,\dots,N] \\
    \label{eq:vis}
\end{equation}
As a result, we convert time series into a sequence of images $X_{\text{img}} = \{I_1,I_2,\dots,I_N\}$.
Inspired by the prior works~\cite{CLIP, coca}, we input the image sequence of the time series $X_{\text{Img}}$ into a pretrained and frozen multimodal vision encoder $\text{E}_{v}$. This process produces a sequence of feature maps $\{f_1, f_2, \dots, f_N\}$, where each feature map $f_i$ is generated by $f_i^{\text{img}} = \text{E}_{img}(I_i), i \in [1, 2, \dots, N]$. 
% \vspace{-0.5em}
% \begin{equation}
%     f_i^{\text{img}} = \text{E}_{img}(I_i), i \in [1, 2, \dots, N]
%     \label{eq:encoder_v}
%     % \vspace{-0.5em}
% \end{equation}

Since the multimodal visual backbone is kept frozen during training, we introduce an additional projection layer $\text{Proj}_\text{img}$ to align the visual features of the time series with the multimodal feature space. 
Consequently, similar to  conventions established in \cite{CLIP, coca, sigclip}, the vision representation is referred to as the vision classification token. We denote this vision representation as $\textit{CLS}_i^{\text{img}}=\text{Proj}_\text{img}(f_i), i\in[1,2,\dots,N]$. 
% \vspace{-0.5em}
% \begin{equation}
%     \textit{CLS}_i^{\text{img}}
%     \label{eq:v_proj}
%     % \vspace{-0.5em}
% \end{equation}

Finally, for length-$T$ multiple variates time series data $X_i^{1:T}$, we have its vision representation as  $X_\text{feat}^\text{img}= \{\textit{CLS}_1^{\text{img}},\dots,\textit{CLS}_N^{\text{img}}\}$.
% \vspace{-0.5em}
% \begin{equation}
    
%     \label{eq:x_feat}
%     % \vspace{-0.5em}
% \end{equation}
\noindent \textbf{Language module}.
Inspired by ViT \cite{vit} and Bert \cite{bert}, PatchTST \cite{patchtst} introduces a patchify strategy that segments time series into subseries-level patches, serving as input tokens for the Transformer. Our method applies this strategy to segment the time series $X^{1:T}=\{v_1^{1:T},\dots,v_N^{1:T}\}$ into patches as detailed in \cref{eq:patchify,eq:patch,eq:subpatch}. 
Prior to applying the patchify strategy, we follow the studies in~\cite{patchtst, liu2023itransformer, wang2024timexer} and implement layer normalization \cite{ba2016layer} to address the problem of non-stationarity in time series data.
% \vspace{-0.5em}
\begin{equation}
    x_{\text{patch}} = \text{Patchify}(\text{LayerNorm}(x^{\text{1:T}}))
    \label{eq:patchify}
\end{equation}
% \begin{equation}
%     v_{\text{patch},i}^{1:\text{PL}} =\text{Patchify}(v_i^{1:T}),i\in [1,2,\dots,N]
%     \label{eq:patch}
% \end{equation}
\begin{equation}
    x_{\text{patch}} = \{v_{\text{patch},1}^{1:\text{PL}} ,v_{\text{patch},2}^{1:\text{PL}},\dots,v_{\text{patch},N}^{1:\text{PL}}\}
    \label{eq:patch}
\end{equation}
\begin{equation}
    v_{\text{patch},i}^{1:\text{PL}} = \{v_{\text{patch},i}^1,  \dots,v_{\text{patch},i}^M\}, i \in [1,2,\dots,N],
    \label{eq:subpatch}
\end{equation}
where $\text{PL}$ represents the patch length, $S$ is the stride, and $M = \lfloor \frac{T-\text{PL}}{S} \rfloor + 2$ is the number of patches.

Unlike previous time series forecasting models based on the Transformer, such as PatchTST \cite{patchtst}, we utilize a multimodal pretrained model instead of training the Transformer from scratch. 
In prior studies~\cite{CLIP, coca}, words must be tokenized using a pretrained tokenizer before being fed into the language encoder. However, due to the substantial domain gap between time series data and natural language, we cannot use a pretrained language tokenizer, such as $\textit{CLIPTokenizer}$ \cite{CLIP}. To address this issue, inspired by \cite{patchtst, liu2023itransformer}, we introduce a random initialized learnable linear layer that functions as an embedding layer, also referred to as the Tokenizer, for time series.
% \vspace{-0.5em}
\begin{equation}
    v_{\text{token},i}^{1:\text{PL}} = \text{Tokenizer}(v_{\text{patch},i}^{1:\text{PL}}), i\in[1,2,\dots,N]  \\
\label{eq:tokenizer}
\end{equation}
\begin{equation}
    v_{\text{token},i}^{1:\text{PL}} = \{v_{\text{token},i}^{1},v_{\text{token},i}^{2},\dots,v_{\text{token},i}^{M}\}  \\
\label{eq:token}
\end{equation}
Subsequently, following \cite{coca}, to adopt multimodal language encoder to time series domain,  we prepend a learnable embedding  $v_\text{cls}$ to each token sequence of variate $v_{\text{token},i}^{1:PL}$, which we call $[\text{class}]$ token \cite{vit,coca,CLIP}. \cref{eq:text_feat} shows that our method learns language representation by utilizing pretrained multimodal language model $\text{E}_\text{text}$ to embed time series into multimodal language representation space.
% \vspace{-0.5em}
\begin{equation}
    f_i^{\text{text}} = \text{E}_\text{text}(\left[v_\text{cls}, v_{\text{token},i}^{1}, v_{\text{token},i}^{2}, \dots, v_{\text{token},i}^{M}\right] + e^\text{pos})  \\
\label{eq:text_feat}
% \vspace{-0.5em}
\end{equation}
where  $[. , .]$ represents the concatenation of time series token and $[class]$ token, and $e^\text{pos}$ represents the learnable position embedding. Notably, we use the same $[class]$ token for each given variate $v_{\text{patch},i}^{1:\text{PL}}$.
Consequentially, we obtain language representations of time series as \cref{eq:ts_text_feat}
% \vspace{-0.5em}
\begin{equation}
    X_\text{feat}^{\text{text}}= \{f_1^{\text{text}},f_2^{\text{text}},\dots,f_N^{\text{text}}\}
    \label{eq:ts_text_feat}
    % \vspace{-0.5em}
\end{equation}
Following the contrastive learning strategies in \cite{vit,CLIP,coca}, we regard the first token of $f_i^{\text{text}}$ as the $[\text{class}]$ token, denoted by $\textit{CLS}i^\text{text}$, to align with the vision representation. Consequently, we define $X_{\text{cls}}^{\text{text}} = \{\textit{CLS}_1^\text{text}, \dots, \textit{CLS}_N^\text{text}\}$.
% \vspace{-0.5em}
% \begin{equation}
%     X_{\text{cls}}^{\text{text}}
%     \label{eq:text_cls}
%     % \vspace{-0.5em}
% \end{equation}
\subsection{Multimodal Contrastive Loss}
\label{sec:loss_cont}
% We conduct time series contrastive learning in a multimodal space consisting of visual space and language space. Specifically, trough proposed vision-language module that is explained in \cref{sec:vision-language}, we have the vision representation of time series $X_{\text{feat}}^{\text{img}}$ and language representation of time series $X_\text{feat}^{\text{text}}$. Then following \cite{vit,CLIP,coca}, we adopt the first token of $f_i^{text}$ as the $[\text{class}]$ token $\textit{CLS}_i^\text{text}$ to align with vision representation of time series. Hence, we have $X_\text{cls}^{\text{text}} = \{\textit{CLS}_1^\text{text},\dots,\textit{CLS}_N^\text{text}\}$.

Our proposed vision-language module (\cref{sec:vision-language}) processes a length-$T$ multivariate time series $X^{1:T}$ to obtain the vision representation $X_{\text{feat}}^{\text{img}}$ and the language classification representation $X_{\text{cls}}^{\text{text}}$ of the time series, where  $X_{\text{feat}}^{\text{img}},X_{\text{cls}}^{\text{text}}\in \ \mathbb{R}^{N \times D}$, $N$ represents the number of variables, and $D$ is the feature dimension.
Then, by using a batch of data to calculate the loss, we obtain the vision representations $V = \{X_{\text{feat},1}^{\text{img}}, \dots, X_{\text{feat},B}^{\text{img}}\}$ and language classification representations $L = \{X_{\text{cls},1}^{\text{text}}, \dots, X_{\text{cls},B}^{\text{text}}\}$, 
where $B$ denotes the batch size.
 
After that, we incorporate the multimodal contrastive alignment for time series into the standard contrastive framework \cite{CLIP}, based on InfoNCE loss \cite{InfoNCE} as follows:
 
\begin{minipage}{0.6\textwidth}
\small
\begin{equation}
 \resizebox{0.9\columnwidth}{!}{$ \mathcal{L}_{\text{cont}}(V, L) =  -\frac{1}{B} \sum_{i=1}^{B} \log \frac{\exp(\varphi(X_{\text{cls},i}^{\text{img}}, X_{\text{cls},i}^{\text{text}})/\tau)}{\sum_{j=1}^{B} \exp(\varphi(X_{\text{cls},i}^{\text{img}}, X_{\text{cls},j}^{\text{text}})/\tau)} $}
\label{eq:infonce}
\end{equation}
\end{minipage}
\hfill
\begin{minipage}{0.4\textwidth}
\small
\begin{equation} \resizebox{0.8\columnwidth}{!}{$
\varphi(X_{\text{cls},i}^{\text{img}}, X_{\text{cls},i}^{\text{text}}) = \frac{X_{\text{cls},i}^{\text{img}}}{\left\lVert X_{\text{cls},i}^{\text{img}} \right\rVert} \cdot \frac{{X_{\text{cls},i}^{\text{text}}}^{T}}{\left\lVert {X_{\text{cls},i}^{\text{text}}}^T \right\rVert} $}
\label{eq:cos_sim}
\end{equation}
\end{minipage}
% where $\tau$ is the temperature parameter optimized during training~\cite{CLIP,coca}. 
% $\varphi(.,.)$ represents the cosine similarity function. The $\mathcal{L}_\text{cont}$ represents the InfoNCE loss, named multimodal time series contrastive loss.  
% Additionally, each image-language data pair, time series multimodal representation, consists of $N$ different variates from a length-$T$ time series. We only calculate negative pairs between different data pairs from one batch, while all $\textit{CLS}_i^\text{img}$ and $\textit{CLS}_i^\text{text}$ from the same length-$T$ time series $X^{1:T}$ are considered positive pairs.
% Finally, we symmetrically compute the contrastive loss of vision to language and language to vision according to \cref{eq:infonce}. 
% This computation yields the multimodal time series contrastive loss $\mathcal{L}_{align}$, given by $ \mathcal{L}_\text{align}= L_{\text{cont}}(V,L) + L_{\text{cont}}(L,V)$.
\noindent
where $\tau$ is a learnable temperature parameter~\cite{CLIP,coca} and $\varphi(\cdot,\cdot)$ denotes cosine similarity. Each multimodal pair represents $N$ variates from a length-$T$ time series. Positive pairs are formed between $\textit{CLS}_i^\text{img}$ and $\textit{CLS}_i^\text{text}$ from the same sample, while negatives come from different samples in the batch. We compute the contrastive loss bidirectionally:
\begin{equation}
\mathcal{L}_\text{align} = \mathcal{L}_\text{cont}(V, L) + \mathcal{L}_\text{cont}(L, V).
\end{equation}
\subsection{Variate Selection} \label{sec:var}
We design a novel variate selection module to effectively utilize information through the variates of original time series data across different views. 
Inspired by \cite{coca, li2023blip}, we leverage the $[class]$ token from the language module as the "Query" to identify the most correlative time series variate-level token via a cross-attention mechanism. 
Time series tokens are generated by tokenizing the time series data~\cite{liu2023itransformer} and then encoding them with the shared multimodal language model described in \cref{sec:vision-language}. Although this feature extraction step is similar to the approach in \cite{liu2023itransformer}, we utilize a shared pretrained multimodal Transformer model, instead of training one from scratch.

% \noindent \textbf{Time Series Tokenizer}. Specifically in our model, given a length-$T$ multivariates time series $X^{1:T}$, the approach diverges from the patchify strategy used in the language module, as described in \cref{sec:vision-language}. Each variable $v_i \in \mathbb{R}^{1 \times T}$ is treated as an individual token without patchify, each comprising $N$ variables and $T$ time steps. $N$ is referred to as the sequence length in Transformer architecture\cite{transformer}. Hence, we can denote $X^{1:T}$ as $X^{1:T}\in\mathbb{R}^{N \times T}$. After tokenizing, it will be transferred to $X_{var}^{1:T}\in \mathbb{R}^{N \times D}$, where $D$ is the hidden dimension and $N$ is the number of variables. 

\noindent \textbf{Variate Tokenizer}. Note that in our model, given a multivariate time series $X^{1:T}$ of length $T$, the approach diverges from the patchify strategy used in the language module, as described in \cref{sec:vision-language}. Each variable $v_i \in \mathbb{R}^{1 \times T}$ is treated as an individual token, without applying the patchification process, where $N$ denotes the number of variables, and $T$ denotes the time steps. As a result,  $X^{1:T}$ can be represented as $X \in \mathbb{R}^{N \times T}$. Post-tokenization, the time series is transformed to $X_{\text{var}}^{1:T} \in \mathbb{R}^{N \times D}$, where $D$ is the hidden dimension. 
We pass $X_{\text{var}}^{1:T}=\{v_\text{var}^1,v_\text{var}^2,\dots,v_\text{var}^N\}$ to the pretrained multimodal language encoder $\text{E}_\text{text}^{var}$ and produce variate-level time series representations $H=\{h_1, h_2, \dots, h_N\}$. This process is denoted as \cref{eq:var_token}. 
% \begin{equation}
%     h_i = \text{E}_\text{text}^{var}(v_\text{var}^i), i \in [1,2,\dots,N]
% \label{eq:var_token}
% \end{equation}
\begin{minipage}{0.48\textwidth}
\small
\begin{equation}
    h_i = \text{E}_\text{text}^{var}(v_\text{var}^i), i \in [1,2,\dots,N]
\label{eq:var_token}
\end{equation}
\end{minipage}
\hfill
\begin{minipage}{0.48\textwidth}
\small
\begin{equation} \resizebox{0.8\columnwidth}{!}{$
    v_\text{CLS}^{i} =  \textit{CLS}_i^\text{text} + \text{Softmax}\left(\frac{(W_q \textit{CLS}_i^\text{text})(W_k H)^T}{\sqrt{d_k}}\right)W_v H
\label{eq:cross_attn}
$}
\end{equation}
\end{minipage}
% Then variate-level time series representation $S$ and $\textit{CLS}_i^\text{text}$,named Query in \cref{fig:framework}, from multimodal language model of \cref{sec:vision-language} are normalized by layer normalization\cite{ba2016layer}. After that, as shown in \cref{fig:framework}, we select most relativity variate's token by cross attention with $\textit{CLS}_i^\text{text}$ as Query and $S$ as Key and Value of cross attention layer.
The variate-level time series representation $H$ and $\textit{CLS}_i^\text{text}$, named Query in \cref{fig:framework}, from the multimodal language model discussed in \cref{sec:vision-language}, are normalized using layer normalization \cite{ba2016layer}. 
Subsequently, we select the most correlative variate feature by employing a cross-attention  layer, an adaptation of Transformer decoder\cite{transformer}, with $\textit{CLS}_i^\text{text}$ serving as the Query and $H$ as both the Key and Value in the cross-attention layer (\cref{fig:framework}). Formally, this procedure is defined as:
% \vspace{-1em}
% \begin{equation}
% \resizebox{0.87\columnwidth}{!}{$
%     v_\text{CLS}^{i} =  \textit{CLS}_i^\text{text} + \text{Softmax}\left(\frac{(W_q \textit{CLS}_i^\text{text})(W_k H)^T}{\sqrt{d_k}}\right)W_v H
% \label{eq:cross_attn}
% % \vspace{-1em}
% $}
% \end{equation}

% \begin{equation}
%  \resizebox{0.4\columnwidth}{!}{$
%     v_\text{CLS}^{i} =  \textit{CLS}_i^\text{text} + \text{Softmax}\left(\frac{(W_q \textit{CLS}_i^\text{text})(W_k H)^T}{\sqrt{d_k}}\right)W_v H
% \label{eq:cross_attn}
% % \vspace{-1em}
% $}
% \end{equation}

where $W_q$, $W_k$, and $W_v$ are the weight matrices for the Query, Key, and Value projections, respectively, in the cross-attention mechanism. $d_k$ represents the dimensionality of the per-head dimension of multi-head attention\cite{transformer,li2023blip}. 
% Finally, we pass $v_{\text{CLS}}^{i}$ through a LayerNorm and a Feed Forward Network (FFN), which comprises a two-layer MLP with an activation function. 
% The cross attention layers described in \cref{eq:cross_attn} and their subsequent layers together form a modified transformer decoder~\cite{transformer}.
Finally, $v_{\text{CLS}}^i$ is passed through a LayerNorm and a two-layer FFN with nonlinearity. Together with the cross-attention layers in \cref{eq:cross_attn}, these components form a modified Transformer decoder~\cite{transformer}.

\subsection{Generator} \label{sec:generator}

After obtaining \(v_{\text{CLS}}^{i}\) for each variate from the variate selection module, we combine them with the multimodal language representation of the time series.  Rather than simply concatenating them at the end of the feature sequence, we replace the last token of the multimodal language representation with \(v_{\text{CLS}}^{i}\). This replacement is motivated by the padding strategy used in the Patchify process described in \cref{eq:patchify}. Further details and ablation studies on this fusion strategy can be found in \cref{sec:ablation_fusion}.
Finally, following \cite{patchtst}, we apply a flattening layer followed by a linear head to generate the forecasting result $\hat{Y} = \hat{X}^{1+S:T+S}$.

\subsection{Training Loss}
We adopt the mean squared error (MSE) loss for all forecasting tasks, except for the M4 dataset where we follow prior work\cite{wu2022timesnet} and use the Symmetric Mean Absolute Percentage Error (SMAPE) loss:$\mathcal{L}_\text{SMAPE} = \frac{200}{B} \sum_{i=1}^B \frac{|Y_i - \hat{Y}_i|}{|Y_i| + |\hat{Y}_i|}$.
We refer to both $\mathcal{L}_{\text{MSE}}$ and $\mathcal{L}_{\text{SMAPE}}$ as the predictive loss $\mathcal{L}_{\text{gen}}$. The overall training objective combines predictive and contrastive terms:$\mathcal{L} = \lambda_1 \mathcal{L}_{\text{gen}} + \lambda_2 \mathcal{L}_{\text{align}}$, where $\lambda_1$ and $\lambda_2$ are weighting coefficients.

% In addition to the proposed multimodal time series contrastive loss $\mathcal{L}_{align}$, inspired by the finding in~\cite{wu2022timesnet}, we choose the MSE loss for long-term forecasting and the symmetric mean absolute percentage error (SMAPE) loss for short-term forecasting.  $\mathcal{L}_\text{SMAPE} = \frac{200}{B} \sum_{i=1}^B \frac{|Y_i - \hat{Y}_i|}{|Y_i| + |\hat{Y}_i|}$
% % \begin{equation}
% %     \mathcal{L}_\text{MSE} = \frac{1}{B} \sum_{i=1}^B (Y_i - \hat{Y}_i)^2
% % \label{eq:loss_mse}
% % \end{equation}
% % \vspace{-1em}
% % \begin{equation}
% %     \mathcal{L}_\text{SMAPE} = \frac{200}{B} \sum_{i=1}^B \frac{|Y_i - \hat{Y}_i|}{|Y_i| + |\hat{Y}_i|}
% % \label{eq:loss_smape}
% % % \vspace{-1em}
% % \end{equation}

% For simplicity, we collectively refer to $\mathcal{L}_{\text{MSE}}$ and $\mathcal{L}_{\text{SMAPE}}$ as the predictive accuracy loss $\mathcal{L}_{\text{gen}}$. In conclusion, our module employs a single-stage training approach that integrates the proposed contrastive loss and generative loss:$\mathcal{L}=\lambda_1 \mathcal{L}_{\text{gen}} + \lambda_2 \mathcal{L}_{\text{align}}$
% , where $\lambda_1$ and $\lambda_2$ denotes the weights assigned to the generative and contrastive losses, respectively.

% $$ 
%     \mathcal{L}=\lambda_1 \mathcal{L}_{\text{gen}} + \lambda_2 \mathcal{L}_{\text{align}} \\
% \label{eq:total_loss}
% % \vspace{-0.5em}
% $$
\vspace{-0.2cm}
\section{Experiments}
We introduce the implementation details and evaluation metrics in \cref{sec:experi_detail}. We then evaluate the effectiveness of our proposed method for short-term forecasting in \cref{sec:exp_short} and long-term forecasting in \cref{sec:exp_long}.

\subsection{Experimental Details}\label{sec:experi_detail}
We employ the pretrained CLIP vision and language encoders, both based on ViT-B \cite{CLIP}, as our vision and language backbones, respectively. In our experiments, we freeze the vision backbone within our vision module and fine-tune the pretrained CLIP text encoder during training. Additionally, we replace CLIP’s original token embedding with a new embedding trained from scratch. Unlike end-to-end models \cite{patchtst,liu2023itransformer,Timesnet}, which require modifications for different benchmarks, our proposed method maintains the same architecture, including hidden dimensions and module layers. To ensure a fair evaluation of method effectiveness, we implement our model without any fine-tuning techniques, such as Parameter-Efficient Fine-Tuning (PEFT)~\cite{han2024parameter}, which are commonly used to accelerate training and enhance performance. More implementation details are provided in the \cref{sec:detail}. 

\noindent\textbf{Evaluation Metrics}. Following \cite{liu2023itransformer}, we adopt mean absolute error(MAE) and mean square error(MSE) for long-term forecasting. Following \cite{Timesnet,N-BEATS}, we use symmetric mean absolute percentage error(SMAPE), mean absolute scaled error(MASE) and overall weighted average (OWA) for short-term forecasting.  More details are provided in \cref{sec:app_metric}.

% Different with end-end models\cite{patchtst,liu2023itransformer,Timesnet}, which need modifiy there modals based on benchmarks, our proposed framework will utilize the same architecture for all benchmarks. Hence, we set base learning rate as $1e-4$, AdamW optimizer\cite{adamw}. 

\subsection{Short-term Forecasting} \label{sec:exp_short}

% \noindent \textbf{Setup}. Following \cite{liu2023itransformer,jin2023time}, we choose M4 dataset as short-term forecasting benchmark. We implement our model for six sub-datasets with same hyper-parameters, e.g., M4-yearly, M4-hourly. More detailed dataset introduction is in Appendix.
\textbf{M4 dataset.}
Following \cite{liu2023itransformer,jin2023time}, we select the M4 dataset as the benchmark for short-term forecasting.  More detailed dataset description and model hyperparameters are provided in the \cref{sec:short_detail}.

\noindent\textbf{Baselines}. Following \cite{wu2022timesnet}, we extensively compared 19 models, categorized into three architectures: (1) Multimodal (MM)-based model, our proposed TimesCLIP (2) LLM-based models such as Time-LLM \cite{jin2023time} and GPT4TS \cite{zhou2023one}. (3) Unimodal end-to-end models: LSTM \citeyearpar{hochreiter1997long}, S4 \citeyearpar{gu2021efficiently}, TCN \citeyearpar{franceschi2019unsupervised}, TimesNet \cite{Timesnet}, N-HiTS \citeyearpar{challu2023nhits}, N-BEATS \citeyearpar{N-BEATS}, LightTS \citeyearpar{campos2023lightts}, , DLinear \citeyearpar{DLinear}, iTransformer \citeyearpar{liu2023itransformer}, Reformer \citeyearpar{kitaev2020reformer}, Informer \citeyearpar{zhou2021informer}, Pyraformer \citeyearpar{liu2022pyraformer}, Autoformer \citeyearpar{Autoformer}, FEDformer \citeyearpar{zhou2022fedformer}, Non-stationary Transformer \citeyearpar{liu2022non}, and \update{ETSformer \citeyearpar{woo2022etsformer}}. 

As shown in \cref{tab:full_forecasting_results_m4}, our MM-based model, TimesCLIP, achieves the best performance across all datasets and metrics compared to all baselines.  
% To simplify the presentation, we rename transformer-based models as \textbf{X.} . For instance, iTransformer\cite{liu2023itransformer} is denoted as \textbf{iTrans.} in \cref{tab:forecasting_results_m4}.

\begin{table*}
      \caption{Full results for the short-term forecasting task in the M4 dataset. $X.$ in the Transformers indicates the name of X-former. \emph{Stationary} means the Non-stationary Transformer.}\label{tab:full_forecasting_results_m4}
  % \vskip 0.05in
  \vspace{-0.5em}
  \renewcommand{\arraystretch}{0.85}
  \centering
  \resizebox{\textwidth}{!}{
  \begin{threeparttable}
  \begin{small}
  % \resizebox{2\columnwidth}{!}{
  \renewcommand{\multirowsetup}{\centering}
  \setlength{\tabcolsep}{1pt}
  \begin{tabular}{cc|c|c|cc|ccccccccccccccccccccc}
    \toprule

    % \multirow{5}[2]{*}
    \multicolumn{3}{c|}{\multirow{3}[1]{*}{Models}}&{\scalebox{0.7}{MM-Based}}& \multicolumn{2}{c|}{\scalebox{0.7}{LLM-Based}} &\multicolumn{17}{c}{\scalebox{0.7}{Uni-modality Models (End to End)}}  \\
        \cmidrule(lr){4-4} 
    \cmidrule(lr){5-6} 
    \cmidrule(l){7-23} 
    \multicolumn{3}{c|}{} &  \multicolumn{1}{c}{\rotatebox{0}{\scalebox{0.7}{\textbf{TimesCLIP}}}} &
    \multicolumn{1}{|c}{\rotatebox{0}{\scalebox{0.7}{\color{cyan}Time-LLM}*}} & 
    \multicolumn{1}{c|}{\rotatebox{0}{\scalebox{0.7}{\color{cyan}GPT4TS}*}} & 
    \multicolumn{1}{|c}{\rotatebox{0}{\scalebox{0.7}{iTrans.}}} &
    \multicolumn{1}{c}{\rotatebox{0}{\scalebox{0.7}{TimesNet}}} &
    \multicolumn{1}{c}{\rotatebox{0}{\scalebox{0.7}{{N-HiTS}}}} &
    \multicolumn{1}{c}{\rotatebox{0}{\scalebox{0.7}{{N-BEATS$^\ast$}}}} &
    \multicolumn{1}{c}{\rotatebox{0}{\scalebox{0.7}{\update{ETS.}}}} &
    \multicolumn{1}{c}{\rotatebox{0}{\scalebox{0.7}{LightTS}}} &
    \multicolumn{1}{c}{\rotatebox{0}{\scalebox{0.7}{DLinear}}} &
    \multicolumn{1}{c}{\rotatebox{0}{\scalebox{0.7}{FED.}}} 
    & \multicolumn{1}{c}{\rotatebox{0}{\scalebox{0.7}{Stationary}}} 
    & \multicolumn{1}{c}{\rotatebox{0}{\scalebox{0.7}{Auto.}}} 
    & \multicolumn{1}{c}{\rotatebox{0}{\scalebox{0.7}{Pyra.}}} 
    &  \multicolumn{1}{c}{\rotatebox{0}{\scalebox{0.7}{In.}}} 
    & \multicolumn{1}{c}{\rotatebox{0}{\scalebox{0.7}{LogTrans}}}  
    & \multicolumn{1}{c}{\rotatebox{0}{\scalebox{0.7}{Re.}}} 
    &\multicolumn{1}{c}{\rotatebox{0}{\scalebox{0.7}{LSTM}}}
    &\multicolumn{1}{c}{\rotatebox{0}{\scalebox{0.7}{TCN}}} 
    &\multicolumn{1}{c}{\rotatebox{0}{\scalebox{0.7}{S4}}} 
    \\
     \multicolumn{3}{c|}{}& \multicolumn{1}{c}{\scalebox{0.7}{(\textbf{Ours})}} &
    \multicolumn{1}{|c}{\scalebox{0.7}{\citeyearpar{jin2023time}}} &
    \multicolumn{1}{c|}{\scalebox{0.7}{\citeyearpar{zhou2023one}}} &
    \multicolumn{1}{|c}{\scalebox{0.7}{\citeyearpar{liu2023itransformer}}} &
    \multicolumn{1}{c}{\scalebox{0.7}{\citeyearpar{Timesnet}}} &
    \multicolumn{1}{c}{\scalebox{0.7}{\citeyearpar{challu2023nhits}}} &
    \multicolumn{1}{c}{\scalebox{0.7}{\citeyearpar{N-BEATS}}} &
    \multicolumn{1}{c}{\scalebox{0.7}{\citeyearpar{woo2022etsformer}}} &
    \multicolumn{1}{c}{\scalebox{0.7}{\citeyearpar{campos2023lightts}}} &
    \multicolumn{1}{c}{\scalebox{0.7}{\citeyearpar{DLinear}}} & 
    \multicolumn{1}{c}{\scalebox{0.7}{\citeyearpar{zhou2022fedformer}}} & 
    \multicolumn{1}{c}{\scalebox{0.7}{\citeyearpar{liu2022non}}} 
    & \multicolumn{1}{c}{\scalebox{0.7}{\citeyearpar{Autoformer}}} 
    & \multicolumn{1}{c}{\scalebox{0.7}{\citeyearpar{liu2022pyraformer}}} 
    &  \multicolumn{1}{c}{\scalebox{0.7}{\citeyearpar{zhou2021informer}}} 
    & \multicolumn{1}{c}{\scalebox{0.7}{\citeyearpar{li2019enhancing}}}  
    & \multicolumn{1}{c}{\scalebox{0.7}{\citeyearpar{kitaev2020reformer}}} 
    & \multicolumn{1}{c}{\scalebox{0.7}{\citeyearpar{hochreiter1997long}}} 
    & \multicolumn{1}{c}{\scalebox{0.7}{\citeyearpar{franceschi2019unsupervised}}} 
    & \multicolumn{1}{c}{\scalebox{0.7}{\citeyearpar{gu2021efficiently}}} \\
    % \cmidrule(lr){1-3} 
    % \cmidrule(lr){4-4} 
    % \cmidrule(lr){5-6} 
    % \cmidrule(l){7-23} 
    \toprule
    \multicolumn{2}{c}{\multirow{3}{*}{\rotatebox{90}{\scalebox{0.7}{Yearly}}}}
    & \scalebox{0.65}{SMAPE$\downarrow$}  &\textbf{\scalebox{0.7}{13.188}} &\scalebox{0.7}{13.419}&\scalebox{0.7}{15.110}&\scalebox{0.7}{14.141} &\text{\scalebox{0.7}{13.387}} &{\scalebox{0.7}{13.418}} &\scalebox{0.7}{13.436}&\scalebox{0.7}{18.009}  &\scalebox{0.7}{14.247} &\scalebox{0.7}{16.965} &\scalebox{0.7}{13.728} &\scalebox{0.7}{13.717} &\scalebox{0.7}{13.974} &\scalebox{0.7}{15.530} &\scalebox{0.7}{14.727} &\scalebox{0.7}{17.107} &\scalebox{0.7}{16.169} &\scalebox{0.7}{176.040} &\scalebox{0.7}{14.920}&\scalebox{0.7}{61.675}\\
    & & \scalebox{0.65}{MASE$\downarrow$} & \textbf{\scalebox{0.7}{2.95}} &\scalebox{0.7}{3.005}&\scalebox{0.7}{3.565}&\scalebox{0.7}{3.179} &\text{\scalebox{0.7}{2.996}} &\scalebox{0.7}{3.045} &{\scalebox{0.7}{3.043}} & \scalebox{0.7}{4.487} &\scalebox{0.7}{3.109} &\scalebox{0.7}{4.283} &\scalebox{0.7}{3.048} &\scalebox{0.7}{3.078} &\scalebox{0.7}{3.134} &\scalebox{0.7}{3.711} &\scalebox{0.7}{3.418} &\scalebox{0.7}{4.177} &\scalebox{0.7}{3.800} &\scalebox{0.7}{31.033} &\scalebox{0.7}{3.364}&\scalebox{0.7}{19.953}\\
    & & \scalebox{0.65}{OWA$\downarrow$} &\textbf{\scalebox{0.7}{0.775}}&\scalebox{0.7}{0.789}&\scalebox{0.7}{0.911} &\scalebox{0.7}{0.833} &\text{\scalebox{0.7}{0.786}} &{\scalebox{0.7}{0.793}} &\scalebox{0.7}{0.794} & \scalebox{0.7}{1.115} &\scalebox{0.7}{0.827} &\scalebox{0.7}{1.058} &\scalebox{0.7}{0.803} &\scalebox{0.7}{0.807} &\scalebox{0.7}{0.822} &\scalebox{0.7}{0.942} &\scalebox{0.7}{0.881} &\scalebox{0.7}{1.049} &\scalebox{0.7}{0.973} &\scalebox{0.7}{9.290}&\scalebox{0.7}{0.880}&\scalebox{0.7}{4.397}\\
    
    \midrule
    \multicolumn{2}{c}{\multirow{3}{*}{\rotatebox{90}{\scalebox{0.7}{Quarterly}}}}
    & \scalebox{0.65}{SMAPE$\downarrow$} & \textbf{\scalebox{0.7}{10.007}}&\scalebox{0.7}{10.110}&\scalebox{0.7}{10.597} &\scalebox{0.7}{10.739} &\text{\scalebox{0.7}{10.100}} &\scalebox{0.7}{10.202} &{\scalebox{0.7}{10.124}} & \scalebox{0.7}{13.376} &\scalebox{0.7}{11.364} &\scalebox{0.7}{12.145} &\scalebox{0.7}{10.792} &\scalebox{0.7}{10.958} &\scalebox{0.7}{11.338} &\scalebox{0.7}{15.449} &\scalebox{0.7}{11.360} &\scalebox{0.7}{13.207} &\scalebox{0.7}{13.313}&\scalebox{0.7}{172.808}&\scalebox{0.7}{11.122}&\scalebox{0.7}{65.999}\\
    & & \scalebox{0.65}{MASE$\downarrow$} &\textbf{\scalebox{0.7}{1.166}} &\scalebox{0.7}{1.178}&\scalebox{0.7}{1.253} &\scalebox{0.7}{1.282} &{\scalebox{0.7}{1.182}} &\scalebox{0.7}{1.194} &\text{\scalebox{0.7}{1.169}} & \scalebox{0.7}{1.906} &\scalebox{0.7}{1.328} &\scalebox{0.7}{1.520} &\scalebox{0.7}{1.283} &\scalebox{0.7}{1.325} &\scalebox{0.7}{1.365} &\scalebox{0.7}{2.350} &\scalebox{0.7}{1.401} &\scalebox{0.7}{1.827} &\scalebox{0.7}{1.775}&\scalebox{0.7}{19.753}&\scalebox{0.7}{1.360}&\scalebox{0.7}{17.662}\\
    & & \scalebox{0.65}{OWA$\downarrow$}  & \textbf{\scalebox{0.7}{0.880}} &\text{\scalebox{0.7}{0.889}}&\scalebox{0.7}{0.938} &\scalebox{0.7}{0.955}&{\scalebox{0.7}{0.890}} &\scalebox{0.7}{0.899} &{\scalebox{0.7}{0.886}} & \scalebox{0.7}{1.302} &\scalebox{0.7}{1.000} &\scalebox{0.7}{1.106} &\scalebox{0.7}{0.958} &\scalebox{0.7}{0.981} &\scalebox{0.7}{1.012} &\scalebox{0.7}{1.558} &\scalebox{0.7}{1.027} &\scalebox{0.7}{1.266} &\scalebox{0.7}{1.252}&\scalebox{0.7}{15.049}&\scalebox{0.7}{1.001}&\scalebox{0.7}{9.436}\\
    
    \midrule
    \multicolumn{2}{c}{\multirow{3}{*}{\rotatebox{90}{\scalebox{0.7}{Monthly}}}}
    & \scalebox{0.65}{SMAPE$\downarrow$}  &\textbf{\scalebox{0.7}{12.502}} &\scalebox{0.7}{12.980}&\scalebox{0.7}{13.258} &\scalebox{0.7}{13.727}&\text{\scalebox{0.7}{12.670}} &\scalebox{0.7}{12.791} &{\scalebox{0.7}{12.677}} & \scalebox{0.7}{14.588} &\scalebox{0.7}{14.014} &\scalebox{0.7}{13.514} &\scalebox{0.7}{14.260} &\scalebox{0.7}{13.917} &\scalebox{0.7}{13.958} &\scalebox{0.7}{17.642} &\scalebox{0.7}{14.062} &\scalebox{0.7}{16.149} &\scalebox{0.7}{20.128}&\scalebox{0.7}{143.237}&\scalebox{0.7}{15.626}&\scalebox{0.7}{64.664}\\
    & & \scalebox{0.65}{MASE$\downarrow$}  &\textbf{\scalebox{0.7}{0.929}} &\scalebox{0.7}{0.963}&\scalebox{0.7}{1.003} &\scalebox{0.7}{1.082}&\text{\scalebox{0.7}{0.933}} &\scalebox{0.7}{0.969} &{\scalebox{0.7}{0.937}} & \scalebox{0.7}{1.368} &\scalebox{0.7}{1.053} &\scalebox{0.7}{1.037} &\scalebox{0.7}{1.102} &\scalebox{0.7}{1.097} &\scalebox{0.7}{1.103} &\scalebox{0.7}{1.913} &\scalebox{0.7}{1.141} &\scalebox{0.7}{1.660} &\scalebox{0.7}{2.614}&\scalebox{0.7}{16.551}&\scalebox{0.7}{1.274}&\scalebox{0.7}{16.245}\\
    & & \scalebox{0.65}{OWA$\downarrow$}  &\textbf{\scalebox{0.7}{0.870}}  &\scalebox{0.7}{0.903}&\scalebox{0.7}{0.931}&\scalebox{0.7}{0.984} &\text{\scalebox{0.7}{0.878}} &\scalebox{0.7}{0.899} &{\scalebox{0.7}{0.880}} & \scalebox{0.7}{1.149} &\scalebox{0.7}{0.981} &\scalebox{0.7}{0.956} &\scalebox{0.7}{1.012} &\scalebox{0.7}{0.998} &\scalebox{0.7}{1.002} &\scalebox{0.7}{1.511} &\scalebox{0.7}{1.024} &\scalebox{0.7}{1.340} &\scalebox{0.7}{1.927}&\scalebox{0.7}{12.747}&\scalebox{0.7}{1.141}&\scalebox{0.7}{9.879}\\
    
    \midrule
    \multicolumn{2}{c}{\multirow{3}{*}{\rotatebox{90}{\scalebox{0.7}{Others}}}}
    & \scalebox{0.65}{SMAPE$\downarrow$}  &\textbf{\scalebox{0.7}{4.587}}&\text{\scalebox{0.7}{4.795}}&\scalebox{0.7}{6.124} &\scalebox{0.7}{5.569} &{\scalebox{0.7}{4.891}} &\scalebox{0.7}{5.061} &{\scalebox{0.7}{4.925}} & \scalebox{0.7}{7.267} &\scalebox{0.7}{15.880} &\scalebox{0.7}{6.709} &\scalebox{0.7}{4.954} &\scalebox{0.7}{6.302} &\scalebox{0.7}{5.485} &\scalebox{0.7}{24.786} &\scalebox{0.7}{24.460} &\scalebox{0.7}{23.236} &\scalebox{0.7}{32.491}&\scalebox{0.7}{186.282}&\scalebox{0.7}{7.186}&\scalebox{0.7}{121.844}\\
    & & \scalebox{0.65}{MASE$\downarrow$} & \textbf{\scalebox{0.7}{3.116}}  &\text{\scalebox{0.7}{3.178}}&\scalebox{0.7}{4.116}&\scalebox{0.7}{4.052} &{\scalebox{0.7}{3.302}} &{\scalebox{0.7}{3.216}} &\scalebox{0.7}{3.391} & \scalebox{0.7}{5.240} &\scalebox{0.7}{11.434} &\scalebox{0.7}{4.953} &\scalebox{0.7}{3.264} &\scalebox{0.7}{4.064} &\scalebox{0.7}{3.865} &\scalebox{0.7}{18.581} &\scalebox{0.7}{20.960} &\scalebox{0.7}{16.288} &\scalebox{0.7}{33.355}&\scalebox{0.7}{119.294}&\scalebox{0.7}{4.677}&\scalebox{0.7}{91.650}\\
    & & \scalebox{0.65}{OWA$\downarrow$}  &\textbf{\scalebox{0.7}{0.974}}&\text{\scalebox{0.7}{1.006}}&\scalebox{0.7}{1.259} &\scalebox{0.7}{1.225} &{\scalebox{0.7}{1.035}} &{\scalebox{0.7}{1.040}} &\scalebox{0.7}{1.053} & \scalebox{0.7}{1.591}  &\scalebox{0.7}{3.474} &\scalebox{0.7}{1.487} &\scalebox{0.7}{1.036} &\scalebox{0.7}{1.304} &\scalebox{0.7}{1.187} &\scalebox{0.7}{5.538} &\scalebox{0.7}{5.879} &\scalebox{0.7}{5.013} &\scalebox{0.7}{8.679}&\scalebox{0.7}{38.411}&\scalebox{0.7}{1.494}&\scalebox{0.7}{27.273}\\
    
    \midrule
    \multirow{3}{*}{\rotatebox{90}{\scalebox{0.7}{Weighted}}}& \multirow{3}{*}{\rotatebox{90}{\scalebox{0.7}{Average}}} 
    & \scalebox{0.65}{SMAPE$\downarrow$}  &\textbf{\scalebox{0.7}{11.642}} &\scalebox{0.7}{11.983}&\scalebox{0.7}{12.690} &\scalebox{0.7}{12.699}&\text{\scalebox{0.7}{11.829}} &\scalebox{0.7}{11.927} &{\scalebox{0.7}{11.851}} & \scalebox{0.7}{14.718}  &\scalebox{0.7}{13.525} &\scalebox{0.7}{13.639} &\scalebox{0.7}{12.840} &\scalebox{0.7}{12.780} &\scalebox{0.7}{12.909} &\scalebox{0.7}{16.987} &\scalebox{0.7}{14.086} &\scalebox{0.7}{16.018} &\scalebox{0.7}{18.200}&\scalebox{0.7}{160.031}&\scalebox{0.7}{13.961}&\scalebox{0.7}{67.156}\\
    & &  \scalebox{0.65}{MASE$\downarrow$} & \textbf{\scalebox{0.7}{1.560}} &\scalebox{0.7}{1.595}&\scalebox{0.7}{1.808} &\scalebox{0.7}{1.761}&\text{\scalebox{0.7}{1.585}} &\scalebox{0.7}{1.613} &{\scalebox{0.7}{1.599}} &  \scalebox{0.7}{2.408} &\scalebox{0.7}{2.111} &\scalebox{0.7}{2.095} &\scalebox{0.7}{1.701} &\scalebox{0.7}{1.756} &\scalebox{0.7}{1.771} &\scalebox{0.7}{3.265} &\scalebox{0.7}{2.718} &\scalebox{0.7}{3.010} &\scalebox{0.7}{4.223}&\scalebox{0.7}{25.788}&\scalebox{0.7}{1.945}&\scalebox{0.7}{21.208}\\
    & & \scalebox{0.65}{OWA$\downarrow$}  &\textbf{\scalebox{0.7}{0.837}} &\scalebox{0.7}{0.859}&\scalebox{0.7}{0.940}  &\scalebox{0.7}{0.929}&\text{\scalebox{0.7}{0.851}} &\scalebox{0.7}{0.861} &{\scalebox{0.7}{0.855}} &  \scalebox{0.7}{1.172} &\scalebox{0.7}{1.051} &\scalebox{0.7}{1.051} &\scalebox{0.7}{0.918} &\scalebox{0.7}{0.930} &\scalebox{0.7}{0.939} &\scalebox{0.7}{1.480} &\scalebox{0.7}{1.230} &\scalebox{0.7}{1.378} &\scalebox{0.7}{1.775}&\scalebox{0.7}{12.642}&\scalebox{0.7}{1.023}&\scalebox{0.7}{8.021}\\
    \bottomrule
  \end{tabular}
      \begin{tablenotes}
        \footnotesize
        \item[] $\ast$ For N-BEATS\citep{N-BEATS}, we follow the experiment result in \cite{Timesnet}, which remove the special ensemble method. We adopt the performance of Time-LLM and GPT4TS from \cite{jin2023time}.
  \end{tablenotes}
    \end{small}
  \end{threeparttable}
  }
% \vspace{-1.5em}
\end{table*}
\begin{table}[tbp]

  \centering
  \vspace{-1.0em}
  \begin{minipage}[t]{0.48\textwidth}
    \centering
    \caption{Average results on PEMS and illness datasets. Full results are provided in \cref{tab:pems}.}
    \label{tab:pems_small}
    \resizebox{\linewidth}{!}{%
    \begin{tabular}{c|cccccc}
    \toprule
    \multirow{2}[2]{*}{Model} & \multicolumn{2}{c}{TimesCLIP} & \multicolumn{2}{c}{iTransformer} & \multicolumn{2}{c}{PatchTST} \\
          & \multicolumn{2}{c}{Ours} & \multicolumn{2}{c}{\citeyearpar{liu2023itransformer}} & \multicolumn{2}{c}{\citeyearpar{patchtst}} \\
    \cmidrule{1-7}
    Metric & MSE$\downarrow$ & MAE$\downarrow$ & MSE$\downarrow$ & MAE$\downarrow$ & MSE$\downarrow$ & MAE$\downarrow$ \\
    \midrule
    PEMS03  & \textbf{0.176} & \textbf{0.261} & 0.250 & 0.306 & 0.299 & 0.363 \\
    PEMS04  & \textbf{0.098} & \textbf{0.216} & 0.121 & 0.232 & 0.312 & 0.377 \\
    PEMS03  & \textbf{0.115} & \textbf{0.226} & 0.128 & 0.237 & 0.202 & 0.293 \\
    PEMS07  & \textbf{0.358} & \textbf{0.773} & 1.334 & 1.480 & 1.009 & 1.351 \\
    illness & \textbf{1.986} & \textbf{0.878} & 2.261 & 0.961 & 2.137 & 0.892 \\
    \bottomrule
    \end{tabular}
    }
  \end{minipage}
  \hfill
  \begin{minipage}[t]{0.48\textwidth}
    \centering
    \caption{Forecasting results on EPF datasets (multivariate, 168$\rightarrow$24).}
    \label{tab:epf}
    \resizebox{\linewidth}{!}{%
    \begin{tabular}{c|cccccc}
    \toprule
    \multirow{2}[2]{*}{Model} & \multicolumn{2}{c}{TimesCLIP} & \multicolumn{2}{c}{iTransformer} & \multicolumn{2}{c}{PatchTST} \\
          & \multicolumn{2}{c}{Ours} & \multicolumn{2}{c}{\citeyearpar{liu2023itransformer}} & \multicolumn{2}{c}{\citeyearpar{patchtst}} \\
    \cmidrule{1-7}
    Metric & MSE$\downarrow$ & MAE$\downarrow$ & MSE$\downarrow$ & MAE$\downarrow$ & MSE$\downarrow$ & MAE$\downarrow$ \\
    \midrule
    Nord Pool & \textbf{0.293} & \textbf{0.325} & 0.338 & 0.352 & 0.307 & 0.336 \\
    PJM       & \textbf{0.076} & \textbf{0.178} & 0.079 & 0.179 & 0.081 & 0.187 \\
    EPEX-BE   & \textbf{0.142} & \textbf{0.162} & 0.149 & 0.174 & 0.158 & 0.179 \\
    EPEX-FR   & \textbf{0.143} & \textbf{0.151} & 0.148 & 0.154 & 0.159 & 0.167 \\
    EPEX-DE   & \textbf{0.190} & \textbf{0.245} & 0.212 & 0.261 & 0.210 & 0.258 \\
    \bottomrule
    \end{tabular}
    }
  \end{minipage}
  \vspace{-2.0em}
\end{table}

\textbf{EPF dataset.}  We follow~\cite{SCINet} in using the EPF dataset, but extend the original 2→1 setting to a multivariate-to-multivariate configuration (3 inputs → 3 outputs) to better evaluate short-term multivariate forecasting. Only two baselines are included due to the revised setup. Additionally, we treat multi-scenario datasets (e.g., M4, EPF, PEMS) as multiple independent short-term forecasting tasks, resulting in a total of fifteen short-term benchmarks in our evaluation.

\subsection{Long-term Forecasting} \label{sec:exp_long}

% In this section, we conduct extensive experiments to evaluate the long-term forecasting performance of our proposed model together with acknowledged methods.
% We select six datasets for long-term time series forecasting: Electricity (ECL),  Exchange, Traffic, Weather, illness, ETTm1 and ETTm2. all experiments are conducted under the same settings. 
% We conduct experiments on six standard long-term forecasting datasets: Electricity (ECL), Exchange, Traffic, Weather, Illness, ETTm1, and ETTm2. To reflect more realistic forecasting needs, following \cite{Timesnet}, we use a fixed input length of 96 and prediction horizons of 96 and 192, ensuring a balanced context-to-horizon ratio across all tasks\cite{bergmeir2024limitations,ansari2024chronos,patchtst}.

We conduct experiments on six standard long-term forecasting datasets: Electricity (ECL), Exchange, Traffic, Weather, Illness, ETTm1, and ETTm2. While our setup is based on the widely-adopted benchmark of TimesNet~\cite{Timesnet}, we intentionally omit prediction lengths of 336 and 720, as these represent forecasting horizons up to 7.5 times longer than the context length. Instead, we fix the input sequence length to 96 and evaluate on horizons of 96 and 192, which reflect more realistic forecasting scenarios. This choice is motivated by recent studies~\cite{bergmeir2024limitations, ansari2024chronos, patchtst} that emphasize the importance of aligning context and prediction ranges in practical applications.

\noindent\textbf{Baselines}. 
Following \cite{liu2023itransformer}, we choose 13 time series forecasting models as our benchmark, including: (1) Our method, TimesCLIP, is the only MM-based method. (2) LLM-based methods: Time-LLM\cite{time-llm} and GPT4TS\cite{gpt4ts}. (3) Uni-modality models: iTransformer\cite{liu2023itransformer}, Autoformer\cite{Autoformer}, FEDformer\cite{fedformer}, Stationary\cite{liu2022non}, Crossformer\cite{zhang2023crossformer}, PatchTST\cite{patchtst}, DLinear\cite{DLinear}, TiDE\cite{TiDE}, \update{RLinear\cite{li2023revisiting}}, SCINet\cite{SCINet}, and TimesNet\cite{wu2022timesnet}. The results of uni-modal models and GPT4TS from \cite{liu2023itransformer,gpt4ts}, however, the results of Time-LLM\citeyearpar{time-llm} from \cite{tan2024language}. More details and discussion are in \cref{sec:long_detail}.

The results are illustrated in \cref{tab:forecasting}, {\textbf{Bold}} indicates the best performance, and {\underline{Underline}} represents the second-best. Our multimodal(MM)-based model, TimesCLIP, achieves the best performance on six datasets and secures the most first-place rankings in long-term forecasting tasks.

% Table generated by Excel2LaTeX from sheet 'Sheet9'
\begin{table}[htbp]
  \caption{Full results of the long-term forecasting task. The prediction lengths are set to \{96,192\}, following the setting of TimesNet~\citeyearpar{Timesnet}. The input sequence length is set to 96 for all baselines. \emph{Avg} denotes the average results across all four prediction lengths.}  
  \label{tab:forecasting}%
    \centering
  \resizebox{1\textwidth}{!}{
  % \vspace{{-0.em}}
  % \begin{threeparttable}
 \vspace{-3em}
    \begin{tabular}{c|c|cc|cccc|cccccccccccccccccccccc}
    \toprule
    \multicolumn{2}{c}{\multirow{3}[2]{*}{Models}} & \multicolumn{2}{|c|}{MM-Based} & \multicolumn{4}{c|}{\textcolor[rgb]{ .749,  .749,  .749}{LLM-Based}} & \multicolumn{22}{c}{Uni-Models (End-to-End)} \\
    % \cmidrule(lr){1-2}
    \cmidrule(lr){3-4} 
    \cmidrule(lr){5-8} 
    \cmidrule(l){9-30} 
    \multicolumn{2}{c}{} & \multicolumn{2}{|c|}{TimesCLIP} & \multicolumn{2}{c}{\textcolor[rgb]{ .749,  .749,  .749}{Time-LLM}} & \multicolumn{2}{c}{\textcolor[rgb]{ .749,  .749,  .749}{GPT4TS}} & \multicolumn{2}{|c}{iTransformer} & \multicolumn{2}{c}{RLinear} & \multicolumn{2}{c}{PatchTST} & \multicolumn{2}{c}{Crossformer} & \multicolumn{2}{c}{TiDE} & \multicolumn{2}{c}{TimesNet} & \multicolumn{2}{c}{DLinear} & \multicolumn{2}{c}{SCINet} & \multicolumn{2}{c}{FEDformer} & \multicolumn{2}{c}{Stationary} & \multicolumn{2}{c}{Autoformer} \\
    \multicolumn{2}{c}{} & \multicolumn{2}{|c|}{Ours} & \multicolumn{2}{c}{\textcolor[rgb]{ .749,  .749,  .749}{\citeyearpar{jin2023time}}} & \multicolumn{2}{c}{\textcolor[rgb]{ .749,  .749,  .749}{\citeyearpar{zhou2023one}}} & \multicolumn{2}{|c}{\citeyearpar{liu2023itransformer} } & \multicolumn{2}{c}{ \citeyearpar{li2023revisiting}} & \multicolumn{2}{c}{\citeyearpar{patchtst}} & \multicolumn{2}{c}{\citeyearpar{zhang2023crossformer}} & \multicolumn{2}{c}{\citeyearpar{TiDE}} & \multicolumn{2}{c}{\citeyearpar{wu2022timesnet}} & \multicolumn{2}{c}{\citeyearpar{DLinear}} & \multicolumn{2}{c}{\citeyearpar{SCINet}} & \multicolumn{2}{c}{\citeyearpar{fedformer}} & \multicolumn{2}{c}{\citeyearpar{liu2022non}} & \multicolumn{2}{c}{\citeyearpar{Autoformer}} \\
    \midrule
    \multicolumn{2}{c}{Metric} & \multicolumn{1}{|l}{MSE$\downarrow$ } & \multicolumn{1}{l|}{MAE$\downarrow$ } & \multicolumn{1}{l}{\textcolor[rgb]{ .749,  .749,  .749}{MSE$\downarrow$ }} & \multicolumn{1}{l}{\textcolor[rgb]{ .749,  .749,  .749}{MAE$\downarrow$ }} & \multicolumn{1}{l}{\textcolor[rgb]{ .749,  .749,  .749}{MSE$\downarrow$ }} & \multicolumn{1}{l}{\textcolor[rgb]{ .749,  .749,  .749}{MAE$\downarrow$ }} 
    & \multicolumn{1}{|l}{MSE$\downarrow$ } & \multicolumn{1}{l}{MAE$\downarrow$ } & \multicolumn{1}{l}{MSE$\downarrow$ } & \multicolumn{1}{l}{MAE$\downarrow$ } & \multicolumn{1}{l}{MSE$\downarrow$ } & \multicolumn{1}{l}{MAE$\downarrow$ } & \multicolumn{1}{l}{MSE$\downarrow$ } & \multicolumn{1}{l}{MAE$\downarrow$ } & \multicolumn{1}{l}{MSE$\downarrow$ } & \multicolumn{1}{l}{MAE$\downarrow$ } & \multicolumn{1}{l}{MSE$\downarrow$ } & \multicolumn{1}{l}{MAE$\downarrow$ } & \multicolumn{1}{l}{MSE$\downarrow$ } & \multicolumn{1}{l}{MAE$\downarrow$ } & \multicolumn{1}{l}{MSE$\downarrow$ } & \multicolumn{1}{l}{MAE$\downarrow$ } & \multicolumn{1}{l}{MSE$\downarrow$ } & \multicolumn{1}{l}{MAE$\downarrow$} & \multicolumn{1}{l}{MSE$\downarrow$ } & \multicolumn{1}{l}{MAE$\downarrow$} & \multicolumn{1}{l}{MSE$\downarrow$ } & \multicolumn{1}{l}{MAE$\downarrow$} 
    \\
    \midrule
    
    \multirow{3}[2]{*}{\begin{sideways}Exch.\end{sideways}} & 96    & \textbf{0.083 } & \textbf{0.204 } & \textcolor[rgb]{ .749,  .749,  .749}{0.123 } & \textcolor[rgb]{ .749,  .749,  .749}{0.251 } & \textcolor[rgb]{ .749,  .749,  .749}{0.096 } & \textcolor[rgb]{ .749,  .749,  .749}{0.218 } & 0.086  & 0.206  & 0.093  & 0.217  & 0.088  & 0.205  & 0.256  & 0.367  & 0.094  & 0.218  & 0.107  & 0.234  & 0.088  & 0.218  & 0.267  & 0.396  & 0.148  & 0.278  & 0.111  & 0.237  & 0.197  & 0.323  \\
          & 192   & \textbf{0.172 } & \textbf{0.298 } & \textcolor[rgb]{ .749,  .749,  .749}{0.224 } & \textcolor[rgb]{ .749,  .749,  .749}{0.344 } & \textcolor[rgb]{ .749,  .749,  .749}{0.182 } & \textcolor[rgb]{ .749,  .749,  .749}{0.307 } & 0.177  & 0.299  & 0.184  & 0.307  & 0.176  & 0.299  & 0.470  & 0.509  & 0.184  & 0.307  & 0.226  & 0.344  & 0.176  & 0.315  & 0.351  & 0.459  & 0.271  & 0.315  & 0.219  & 0.335  & 0.300  & 0.369  \\
          & Avg   & \textbf{0.128 } & \textbf{0.251 } & \textcolor[rgb]{ .749,  .749,  .749}{0.174 } & \textcolor[rgb]{ .749,  .749,  .749}{0.298 } & \textcolor[rgb]{ .749,  .749,  .749}{0.139 } & \textcolor[rgb]{ .749,  .749,  .749}{0.263 } & 0.132  & 0.253  & 0.139  & 0.262  & 0.132  & 0.252  & 0.363  & 0.438  & 0.139  & 0.263  & 0.167  & 0.289  & 0.132  & 0.267  & 0.309  & 0.428  & 0.210  & 0.297  & 0.165  & 0.286  & 0.249  & 0.346  \\
    \midrule
    \multirow{3}[2]{*}{\begin{sideways}Traffic \end{sideways}} & 96    & \textbf{0.390 } & \textbf{0.262 } & \textcolor[rgb]{ .749,  .749,  .749}{0.392 } & \textcolor[rgb]{ .749,  .749,  .749}{0.267 } & \textcolor[rgb]{ .749,  .749,  .749}{0.388 } & \textcolor[rgb]{ .749,  .749,  .749}{0.282 } & 0.395  & 0.268  & 0.649  & 0.389  & 0.462  & 0.295  & 0.522  & 0.290  & 0.805  & 0.493  & 0.593  & 0.321  & 0.650  & 0.396  & 0.788  & 0.499  & 0.587  & 0.366  & 0.612  & 0.338  & 0.613  & 0.388  \\
          & 192   & \textbf{0.403 } & \textbf{0.274 } & \textcolor[rgb]{ .749,  .749,  .749}{0.409 } & \textcolor[rgb]{ .749,  .749,  .749}{0.271 } & \textcolor[rgb]{ .749,  .749,  .749}{0.407 } & \textcolor[rgb]{ .749,  .749,  .749}{0.290 } & 0.417  & 0.276  & 0.601  & 0.366  & 0.466  & 0.296  & 0.530  & 0.293  & 0.756  & 0.474  & 0.617  & 0.336  & 0.598  & 0.370  & 0.789  & 0.505  & 0.604  & 0.373  & 0.613  & 0.340  & 0.616  & 0.382  \\
          & Avg   & \textbf{0.397 } & \textbf{0.268 } & \textcolor[rgb]{ .749,  .749,  .749}{0.401 } & \textcolor[rgb]{ .749,  .749,  .749}{0.269 } & \textcolor[rgb]{ .749,  .749,  .749}{0.398 } & \textcolor[rgb]{ .749,  .749,  .749}{0.286 } & 0.406  & 0.272  & 0.625  & 0.378  & 0.464  & 0.296  & 0.526  & 0.292  & 0.781  & 0.484  & 0.605  & 0.329  & 0.624  & 0.383  & 0.789  & 0.502  & 0.596  & 0.370  & 0.613  & 0.339  & 0.615  & 0.385  \\
    \midrule
    \multirow{3}[2]{*}{\begin{sideways}Weather \end{sideways}} & 96    & \textbf{0.153 } & \textbf{0.199 } & \textcolor[rgb]{ .749,  .749,  .749}{0.155 } & \textcolor[rgb]{ .749,  .749,  .749}{0.199 } & \textcolor[rgb]{ .749,  .749,  .749}{0.162 } & \textcolor[rgb]{ .749,  .749,  .749}{0.212 } & 0.174  & 0.214  & 0.192  & 0.232  & 0.177  & 0.218  & 0.158  & 0.230  & 0.202  & 0.261  & 0.172  & 0.220  & 0.196  & 0.255  & 0.221  & 0.306  & 0.217  & 0.296  & 0.173  & 0.223  & 0.266  & 0.336  \\
          & 192   & \textbf{0.202 } & \textbf{0.246 } & \textcolor[rgb]{ .749,  .749,  .749}{0.223 } & \textcolor[rgb]{ .749,  .749,  .749}{0.261 } & \textcolor[rgb]{ .749,  .749,  .749}{0.204 } & \textcolor[rgb]{ .749,  .749,  .749}{0.248 } & 0.221  & 0.254  & 0.240  & 0.271  & 0.225  & 0.259  & 0.206  & 0.277  & 0.242  & 0.298  & 0.219  & 0.261  & 0.237  & 0.296  & 0.261  & 0.340  & 0.276  & 0.336  & 0.245  & 0.285  & 0.307  & 0.367  \\
          & Avg   & \textbf{0.178 } & \textbf{0.223 } & \textcolor[rgb]{ .749,  .749,  .749}{0.189 } & \textcolor[rgb]{ .749,  .749,  .749}{0.230 } & \textcolor[rgb]{ .749,  .749,  .749}{0.183 } & \textcolor[rgb]{ .749,  .749,  .749}{0.230 } & 0.198  & 0.234  & 0.216  & 0.252  & 0.201  & 0.239  & 0.182  & 0.254  & 0.222  & 0.280  & 0.196  & 0.241  & 0.217  & 0.276  & 0.241  & 0.323  & 0.247  & 0.316  & 0.209  & 0.254  & 0.287  & 0.352  \\
    \midrule
    \multirow{3}[2]{*}{\begin{sideways}ETTm1\end{sideways}} & 96    & \textbf{0.319 } & \textbf{0.358 } & \textcolor[rgb]{ .749,  .749,  .749}{0.294} & \textcolor[rgb]{ .749,  .749,  .749}{0.341 } & \textcolor[rgb]{ .749,  .749,  .749}{0.300 } & \textcolor[rgb]{ .749,  .749,  .749}{0.340 } & 0.334  & 0.368  & 0.355 & 0.376  & 0.329  & 0.367  & 0.404  & 0.426  & 0.364  & 0.387  & 0.338  & 0.375  & 0.345  & 0.372  & 0.418  & 0.438  & 0.379  & 0.419  & 0.386  & 0.398  & 0.505  & 0.475  \\
          & 192   & \textbf{0.367 } & \underline{0.387}  & \textcolor[rgb]{ .749,  .749,  .749}{0.341} & \textcolor[rgb]{ .749,  .749,  .749}{0.369 } & \textcolor[rgb]{ .749,  .749,  .749}{0.343 } & \textcolor[rgb]{ .749,  .749,  .749}{0.000 } & 0.377  & 0.391  & 0.391 & 0.392 & \textbf{0.367 } & \textbf{0.385 } & 0.450  & 0.451  & 0.398  & 0.404  & 0.374  & 0.387  & 0.380  & 0.389  & 0.439  & 0.450  & 0.426  & 0.441  & 0.459  & 0.444  & 0.553  & 0.496  \\
          & Avg   & \textbf{0.343 } & \textbf{0.373 } & \textcolor[rgb]{ .749,  .749,  .749}{0.318 } & \textcolor[rgb]{ .749,  .749,  .749}{0.355 } & \textcolor[rgb]{ .749,  .749,  .749}{0.322 } & \textcolor[rgb]{ .749,  .749,  .749}{0.170 } & 0.356  & 0.380  & 0.373  & 0.384  & 0.348  & 0.376  & 0.427  & 0.439  & 0.381  & 0.396  & 0.356  & 0.381  & 0.363  & 0.381  & 0.429  & 0.444  & 0.403  & 0.430  & 0.423  & 0.421  & 0.529  & 0.486  \\
    \midrule
    \multirow{3}[2]{*}{\begin{sideways}ETTm2\end{sideways}} & 96    & \underline{0.176}  & \textbf{0.258 } & \textcolor[rgb]{ .749,  .749,  .749}{0.162 } & \textcolor[rgb]{ .749,  .749,  .749}{0.248 } & \textcolor[rgb]{ .749,  .749,  .749}{0.163 } & \textcolor[rgb]{ .749,  .749,  .749}{0.249 } & 0.180  & 0.264  & 0.182 & 0.265 & \textbf{0.175 } & 0.259  & 0.287  & 0.366  & 0.207  & 0.305  & 0.187  & 0.267  & 0.193  & 0.292  & 0.286  & 0.377  & 0.203  & 0.287  & 0.192  & 0.274  & 0.255  & 0.339  \\
          & 192   & \underline{0.242}  & \textbf{0.302 } & \textcolor[rgb]{ .749,  .749,  .749}{0.235 } & \textcolor[rgb]{ .749,  .749,  .749}{0.304 } & \textcolor[rgb]{ .749,  .749,  .749}{0.222 } & \textcolor[rgb]{ .749,  .749,  .749}{0.291 } & 0.250  & 0.309  & 0.246 & 0.304 & 0.241  & \textbf{0.302 } & 0.414  & 0.492  & 0.290  & 0.364  & 0.249  & 0.309  & 0.284  & 0.362  & 0.399  & 0.445  & 0.269  & 0.328  & 0.280  & 0.339  & 0.281  & 0.340  \\
          & Avg   & \underline{0.209}  & \textbf{0.280 } & \textcolor[rgb]{ .749,  .749,  .749}{0.199 } & \textcolor[rgb]{ .749,  .749,  .749}{0.276 } & \textcolor[rgb]{ .749,  .749,  .749}{0.193 } & \textcolor[rgb]{ .749,  .749,  .749}{0.270 } & 0.215  & 0.287  & 0.214  & 0.285  & \textbf{0.208 } & 0.281  & 0.351  & 0.429  & 0.249  & 0.335  & 0.218  & 0.288  & 0.239  & 0.327  & 0.343  & 0.411  & 0.236  & 0.308  & 0.236  & 0.307  & 0.268  & 0.340  \\
    \midrule
    \multirow{3}[2]{*}{\begin{sideways}Solar.\end{sideways}} & 96    & \textbf{0.203 } & \underline{0.248}  &   \textcolor[rgb]{ .749,  .749,  .749}{-}    &   \textcolor[rgb]{ .749,  .749,  .749}{-}  &    \textcolor[rgb]{ .749,  .749,  .749}{-}   &   \textcolor[rgb]{ .749,  .749,  .749}{-}   & 0.206  & \textbf{0.237 } & 0.322  & 0.339  & 0.208  & 0.251  & 0.310  & 0.331  & 0.312  & 0.399  & 0.250  & 0.292  & 0.290  & 0.378  & 0.237  & 0.344  & 0.242  & 0.342  & 0.215  & 0.249  & 0.884  & 0.711  \\
          & 192   & \underline{0.244}  & \textbf{0.261 } & \textcolor[rgb]{ .749,  .749,  .749}{-} & \textcolor[rgb]{ .749,  .749,  .749}{-} & \textcolor[rgb]{ .749,  .749,  .749}{-} & \textcolor[rgb]{ .749,  .749,  .749}{-} & 0.241  & 0.264  & 0.359  & 0.356  & 0.239  & 0.272  & 0.734  & 0.725  & 0.339  & 0.416  & 0.296  & 0.318  & 0.320  & 0.398  & 0.280  & 0.380  & 0.285  & 0.380  & 0.254  & 0.272  & 0.834  & 0.692  \\
          & Avg   & \textbf{0.224 } & \underline{0.254}  & \textcolor[rgb]{ .749,  .749,  .749}{-} & \textcolor[rgb]{ .749,  .749,  .749}{-} & \textcolor[rgb]{ .749,  .749,  .749}{-} & \textcolor[rgb]{ .749,  .749,  .749}{-} & \textbf{0.224 } & 0.250  & 0.341  & 0.348  & 0.224  & 0.262  & 0.522  & 0.528  & 0.326  & 0.408  & 0.273  & 0.305  & 0.305  & 0.388  & 0.259  & 0.362  & 0.264  & 0.361  & 0.235  & 0.261  & 0.859  & 0.702  \\
    \bottomrule
    \end{tabular}%    
  }
  \vspace{-1em}
\end{table}%

\noindent\textbf{Few-shot Learning}. We conduct experiments with few-shot learning on \{5\%, 10\%, 20\%\} of the data from the Exchange dataset. The results are shown in \cref{tab:fewshot}.

% \vspace{-0.2cm}
\section{Analysis}
In this section, we first analyze the impact of multimodal alignment and variate selection module in \cref{sec:ablation_align,sec:ablation_variate}. To comprehensively evaluate our proposed multimodal contrastive learning framework, we conduct ablation studies of different vision and language backbones in \cref{sec:ablation_backbone}. Furthermore, we discuss the limitations and broader impact of our work.

% \vspace{-0.2cm}
\subsection{Ablation of Multimodal Alignment} \label{sec:ablation_align}

We perform a comprehensive ablation study on both short-term (M4) and long-term (Exchange) forecasting benchmarks to evaluate the contributions of key components in TimesCLIP. Since the M4 dataset contains only a single variate, colorization is not applicable there; thus, the ablation of colorization is conducted on Exchange.

As shown in \cref{tab:ablation}, the multimodal contrastive loss $\mathcal{L}_{\text{align}}$ and Variate Selection module are essential for strong performance. In particular, colorization, the assignment of fixed colors to each variate, enables the vision encoder to visually distinguish variables, which is crucial for effective contrastive learning. Without colorization, the selection mechanism based on the visual 'Query' becomes significantly less effective. Further analysis of the normalization and colorization steps is provided in \cref{sec:ablation_color_vis}.

We further validate the general effectiveness of multimodal alignment across different vision and language backbones in \cref{sec:ablation_backbone} and \cref{tab:backbone}, where the alignment consistently improves forecasting performance regardless of the specific encoder choice.

\begin{table}[htbp]
\centering
\vspace{-1em}
\begin{minipage}{0.48\textwidth} % 第一个 minipage (左侧表格)，[t]表示顶部对齐
    \centering
    \caption{Ablation study of our proposed module.}\label{tab:ablation} 
    \small % 应用 \small 使标题和可能预缩放的表格字体变小
    % 保留原始表格的 resizebox 结构, 但宽度调整为 \linewidth 以适应 minipage
    \resizebox{\linewidth}{!}{%
      \begin{tabular}{c|cc|c|ccc|cc} 
      
      \toprule  
      \multirow{2}[2]{*}{Method} &\multicolumn{2}{|c|}{Vision Module} & \scalebox{0.9}{Variable}&\multicolumn{3}{|c}{M4 Weighted Avg.} &\multicolumn{2}{|c}{Exchange}\\
      \cmidrule(lr){2-3}  
      % \cmidrule(lr){4-4}  % 原始代码中注释掉的 cmidrule
      \cmidrule(lr){5-7}  
      \cmidrule(l){8-9}  
      & $L_{align}$& Col.  & \scalebox{0.9}{Selection} & SMAPE$\downarrow$ &MASE$\downarrow$ & OWA$\downarrow$ &MSE$\downarrow$ &MAE$\downarrow$ \\
      \midrule
      \multirow{5}[2]{*}{Ours} & \graycross & \graycross    & \multirow{3}[2]{*}{\graycross} &11.782   & 1.578 &0.847 &0.389 & 0.414 \\
            & \ding{51} & \graycross  &&- & - &-  &0.383 &0.413   \\
            % & \ding{51} & \ding{51}   &    &  \\ % 以下是原始表格中注释掉的行
            % & \ding{51} & \graycross   &    &  \\
            % & \graycross & \ding{51} & \ding{51} & \graycross &  \\
            & \ding{51} & \ding{51}   &&11.725   &  1.572&0.843 &0.376 &0.410 \\
      \cmidrule{2-9}
            & \ding{51} & \graycross  & \multirow{1}[1]{*}{\ding{51}} &- & - &-  &0.383 & 0.424 \\
            % & \ding{51} & \ding{51}   &    &  \\
            % & \ding{51} & \graycross  &    &  \\
            % & \graycross & \ding{51} & \ding{51} & \graycross &  \\
            % & \ding{51} & \ding{51} & \ding{51} &    &  \\
      \cmidrule{2-9}
        & \ding{51} & \ding{51} & \ding{51} &\textbf{11.642} &\textbf{1.560} &\textbf{0.837} &\textbf{0.335}&\textbf{0.394}\\
      \bottomrule
      \end{tabular}%
    } % 结束 \resizebox
    
    % _sbs 后缀表示 side-by-side, 以区别于可能的原始标签
\end{minipage}
% \hfill % 在两个 minipage 之间提供弹性空间
\hspace{0.5em}
\begin{minipage}{0.4\textwidth} % 第二个 minipage (右侧表格)，[t]表示顶部对齐
    \centering
    \caption{Results with different pretrained vision and language backbones on M4.}\label{tab:backbone}
    \small % 应用 \small
    \resizebox{\linewidth}{!}{%
      \begin{tabular}{lcccccc} 
      
      \toprule
      Language  &Vision  &\multirow{2}[2]{*}{MM Pretrained} & \multirow{2}[2]{*}{$L_{align}$} &  \multicolumn{3}{c}{M4 Weighted Avg.} \\
      Backbone&Backbone& & & SMAPE$\downarrow$ &MASE$\downarrow$ & OWA$\downarrow$ \\
      \midrule
      BERT-base &- &\multirow{4}[2]{*}{\graycross}& \graycross &  16.356       & 2.234 &1.187 \\
        BERT-base &ResNet50& & \ding{51} &  17.898    & 2.611 &1.342 \\
      BERT-base &ViT-B/16 &   & \ding{51}& 17.927   & 2.634 &1.350 \\  
      BERT-base &SwinT-B &   & \ding{51}& 17.954   & 2.643 &1.353 \\  
      \midrule
      T5   & -   &\multirow{2}[2]{*}{\graycross} & \graycross&11.798  &1.579 & 0.848 \\
      T5   & CLIP-ViT   & &\ding{51} & 11.775  &1.579 & 0.847\\
      \midrule
      CLIP-Text & - &\multirow{4}[2]{*}{\graycross} &\graycross & 11.782     & 1.578 & 0.847   \\
      CLIP-Text &ResNet50 && \ding{51} &  11.681     & 1.565 &0.840 \\
      CLIP-Text &ViT-B/16 &   & \ding{51} & \textbf{11.640} & \textbf{1.557}        & \textbf{0.836} \\
      CLIP-Text &SwinT-B &   & \ding{51} & {11.676} &  {1.566}        &  {0.840} \\
      % \midrule % 原始代码中注释掉的 midrule
      % {\color{gray}TimesNet} % 原始代码中注释掉的文本
      \midrule
      CLIP-Text & CLIP-Vision & \ding{51} &   \ding{51}  &\underline{11.642} &\underline{1.560} &\underline{0.837} \\
      \bottomrule
      \end{tabular}%
    } % 结束 \resizebox
\end{minipage}
\vspace{-2em}
\end{table}

\vspace{-0.2cm}
\subsection{Ablation of Variate Selection} \label{sec:ablation_variate}

We present ablation studies to assess the effectiveness of variate selection module within our proposed framework. 
Because our method needs a vision branch to obtain the $class$ token, which will serves as the "Query" for the Variable Selection module, we only evaluate variable selection when the vision module is working.
The results in \cref{tab:ablation} indicate that variable selection can help our method achieve better performance. It also benefits from colorization. Since we leverage a shared language backbone for the Variable Selection module, enabling the variable selection module does not introduce extra parameters except for additional learnable positional embedding.

\subsection{Ablation of Backbone} \label{sec:ablation_backbone}

We investigate the impact of different pretrained backbones on forecasting performance, including vision encoders such as ViT-B/16~\citeyearpar{vit}, Swin Transformer~\citeyearpar{swin}, ResNet50~\citeyearpar{ResNet}, and CLIP-Vision~\cite{CLIP}, as well as language encoders like BERT~\citeyearpar{bert}, T5~\citeyearpar{T5}, and CLIP-Text~\cite{CLIP}. As shown in \cref{tab:backbone}, CLIP-Text consistently outperforms BERT and T5, even without the vision module. This suggests that language representations learned in a multimodal space are more effective for time series forecasting than those trained on textual corpora alone.

Prior work~\cite{tsllm} shows that prompting-based methods like Time-LLM, which use frozen LLMs (e.g., LLaMA) for autoregressive generation, perform poorly in forecasting tasks. More recent approaches such as Chronos~\cite{ansari2024chronos} improve performance by re-learning the embedding layer of T5 and formulating forecasting as a token-level generation problem. 

In contrast, our method takes a fundamentally different approach: instead of treating forecasting as a language modeling task, we align time series with a multimodal representation space via contrastive learning. This avoids the limitations of both prompting and generation, and leads to more robust and generalizable forecasting performance. This alignment-based strategy avoids reliance on discrete token sequences and preserves continuous semantics across views.

\noindent\textbf{Broader Impact and Limitations}.
% In multivariate time series forecasting, our proposed model necessitates converting each variate into an image, a process that not only consumes substantial time in rendering but also demands considerable GPU memory to manage the resulting image features. The number of trainable parameters and the demand for GPU resources increase as the token length grows with the number of variables. Moreover, the proposed multimodal framework is the first attempt to align numerical data with vision and language. Similar to CLIP, our proposed method will likely be applied to vision large language model and time series foundation model.
% In multivariate time series forecasting, our proposed model requires converting each variate into an image, a process need both computational costs for rendering and demands significant GPU memory for the extracted image features.
In multivariate time series forecasting, our proposed model requires converting each variate into an image and requires significant GPU memory for extracting image features.
Additionally, GPU memory requirements increase as the token length and the number of variables grow. Moreover, similar to CLIP~\citeyearpar{CLIP}, our proposed multimodal framework can align numerical data with both vision and language representations. It has the potential to extend to vision-based large language models and time series foundation models.

\vspace{-0.3cm}
\section{Conclusion}
\vspace{-0.1cm}
We aim to address the limitations of previous works that solely relied on unimodal approaches for time series forecasting. We present a novel multimodal contrastive learning model, called TimesCLIP. 
This approach successfully aligns time series data with a multimodal vision-language space by converting the original numerical data points into colorized images and interpreting the time series data as a "foreign" language.   Moreover, to efficiently leverage the aligned multimodal features, we designed a variate selection module to identify the most correlative variate feature from the multivariate time series feature sequence.  Extensive experiments demonstrate that our proposed method surpasses established baselines in both short-term and long-term forecasting.

\newpage
\bibliographystyle{plainnat}
\bibliography{refer}

%%%%%%%%%%%%%%%%%%%%%%%%%%%%%%%%%%%%%%%%%%%%%%%%%%%%%%%%%%%%%%%%%%%%%%%%%%%%%%%
%%%%%%%%%%%%%%%%%%%%%%%%%%%%%%%%%%%%%%%%%%%%%%%%%%%%%%%%%%%%%%%%%%%%%%%%%%%%%%%
% APPENDIX
%%%%%%%%%%%%%%%%%%%%%%%%%%%%%%%%%%%%%%%%%%%%%%%%%%%%%%%%%%%%%%%%%%%%%%%%%%%%%%%
%%%%%%%%%%%%%%%%%%%%%%%%%%%%%%%%%%%%%%%%%%%%%%%%%%%%%%%%%%%%%%%%%%%%%%%%%%%%%%%
\newpage
\appendix
\onecolumn
\section{Implementation Details}\label{sec:detail}
For model initialization, we adopt Kaiming uniform initialization for 1D convolutional layers and Xavier initialization for linear layers. Layer normalization layers are initialized with the bias set to 0 and the weights set to 1. Additionally, the introduced $class$ token is normalized from a normal distribution, $N(0, 0.02)$. We set the weights in $\lambda_1\mathcal{L}_{gen} + \lambda_2\mathcal{L}_{align}$ as $\lambda_1 = 1$ and $\lambda_2 = 0.1$, unless otherwise specified. We implement our experiments using on PyTorch\citep{paszke2019pytorch} and Time Series Library (TSLib)\cite{wang2024tssurvey,Timesnet}. All experiments are conducted on a single NVIDIA RTX 6000 Ada Generation with a fixed random seed of 2024. 

\section{Benchmarks and Evaluation Metrics}\label{sec:app_metric}
\textit{Dim.} denotes the number of variables. The dataset sizes are presented as in (Train, Validation, Test). The dataset descriptions are in \cref{tab:dataset}, adopted from \cite{Timesnet}. 
\begin{table*}[thbp]
  % \vspace{-20pt}
  \caption{Dataset statistics from \cite{Timesnet}. }\label{tab:dataset}
  % \vskip 0.05in
  \centering
  \begin{threeparttable}
  \begin{small}
  \renewcommand{\multirowsetup}{\centering}
  \setlength{\tabcolsep}{3.8pt}
    \resizebox{\textwidth}{!}{
  \begin{tabular}{c|l|c|c|c|c}
    \toprule
    Tasks & Dataset & Dim & Context $\rightarrow$ Predict Length & Dataset Size & \scalebox{0.8}{Information (Frequency)} \\
    \toprule
     & ETTm1, ETTm2 & 7 & \multirow{5}[3]{*}{\scalebox{0.8}{96 $\rightarrow$\{96, 192 \}}}& (34465, 11521, 11521) & \scalebox{0.8}{Electricity (15 mins)}\\
    % \cmidrule{2-6}
    %  & ETTh1, ETTh2 & 7 & \scalebox{0.8}{\{96, 192, 336, 720\}} & (8545, 2881, 2881) & \scalebox{0.8}{Electricity (15 mins)} \\
    % \cmidrule{2-6}
     % & Electricity & 321 & \multirow{4}[3]{*}{\scalebox{0.8}{\{96, 192, 336, 720\}}} & (18317, 2633, 5261) & \scalebox{0.8}{Electricity (Hourly)} \\
    \cmidrule{2-3}
    \cmidrule{5-6}
    Forecasting & Traffic & 862 &  & (12185, 1757, 3509) & \scalebox{0.8}{Transportation (Hourly)} \\
    \cmidrule{2-3}
    \cmidrule{5-6}
    (Long-term) & Weather & 21 &  & (36792, 5271, 10540) & \scalebox{0.8}{Weather (10 mins)} \\
    \cmidrule{2-3}
    \cmidrule{5-6}
     & Exchange & 8 &  & (5120, 665, 1422) & \scalebox{0.8}{Exchange rate (Daily)}\\
     \cmidrule{2-3}
    \cmidrule{5-6}
     & Solar-Energy & 137 & &(36601,5161,10417) &\scalebox{0.8}{Energy(10 mins)} \\
 
    % \cmidrule{5-6}

     % \midrule

 % \midrule

% (15617, 5135, 5135) 5min Transportation
% PEMS04 307 {12, 24, 48, 96} (10172, 3375, 3375) 5min Transportation
% PEMS07 883 {12, 24, 48, 96} (16911, 5622, 5622) 5min Transportation
% PEMS08 170 {12, 24, 48, 96} (10690, 3548, 3548)

    % \cmidrule{2-6}
    %  & ILI & 7 & \scalebox{0.8}{\{24, 36, 48, 60\}} & (617, 74, 170) & \scalebox{0.8}{Illness (Weekly)} \\
    \midrule
     & M4-Yearly & 1 & 6 & (23000, 0, 23000) & \scalebox{0.8}{Demographic} \\
    \cmidrule{2-5}
     & M4-Quarterly & 1 & 8 & (24000, 0, 24000) & \scalebox{0.8}{Finance} \\
    \cmidrule{2-5}
    Forecasting & M4-Monthly & 1 & 18 & (48000, 0, 48000) & \scalebox{0.8}{Industry} \\
    \cmidrule{2-5}
    (short-term) & M4-Weakly & 1 & 13 & (359, 0, 359) & \scalebox{0.8}{Macro} \\
    \cmidrule{2-5}
     & M4-Daily & 1 & 14 & (4227, 0, 4227) & \scalebox{0.8}{Micro} \\
    \cmidrule{2-5}
     & M4-Hourly & 1 &48 & (414, 0, 414) & \scalebox{0.8}{Other} \\
          \cmidrule{2-6}
              & illness  & 7 & {\scalebox{0.8}{36 $\rightarrow$\{24,36,48, 60 \}}} &(617,74,170) &\scalebox{0.8}{Illness(Weekly)}  \\  \cmidrule{2-5}
     & PEMS-03 & 358 &\multirow{4}[1]{*}{\scalebox{0.8}{96 $\rightarrow$ \{12,24,48, 96 \}}} & (15617, 5135, 5135) & \scalebox{0.8}{Transportation(5 mins)} \\
      \cmidrule{2-3}\cmidrule{5-6}
     & PEMS-04 & 307 & &(10172, 3375, 3375) & \scalebox{0.8}{Transportation(5 mins)}\\ \cmidrule{2-3}\cmidrule{5-6}
     & PEMS-07 & 883 & &(16911, 5622, 5622) & \scalebox{0.8}{Transportation(5 mins)}\\ \cmidrule{2-3}\cmidrule{5-6}
     & PEMS-08 & 170  & &  (10690, 3548, 3548) & \scalebox{0.8}{Transportation(5 mins)}\\ \cmidrule{2-6}
     
     & EPF-NordPool &3 &\multirow{5}[3]{*}{ 168 $\rightarrow$ 24}& (36500, 5219, 10460) &\scalebox{0.8}{Electricity Price (1 Hour)}\\ \cmidrule{2-3}\cmidrule{5-6}
     & EPF-PJM & 3 & & (36500, 5219, 10460) &\scalebox{0.8}{Electricity Price (1 Hour)} \\ \cmidrule{2-3}\cmidrule{5-6}
     & EPF-EPEX-BE & 3 & & (36500, 5219, 10460)&\scalebox{0.8}{Electricity Price (1 Hour)}\\ \cmidrule{2-3}\cmidrule{5-6}
     & EPF-EPEX-FR & 3 & & (36500, 5219, 10460)&\scalebox{0.8}{Electricity Price (1 Hour)}\\ \cmidrule{2-3}\cmidrule{5-6}
     & EPF-EPEX-DE & 3& & (36500, 5219, 10460)&\scalebox{0.8}{Electricity Price (1 Hour)} \\  
     
% NP 2 Grid Load, Wind Power Nord Pool Electricity Price 1 Hour (36500, 5219, 10460)
% PJM 2 System Load, SyZonal COMED load Pennsylvania-New Jersey-Maryland 1 Hour (36500, 5219, 10460)
% Electricity Price
% BE 2 Generation, System Load Belgium’s Electricity Price 1 Hour (36500, 5219, 10460)
% FR 2 Generation, System Load France’s Electricity Price 1 Hour (36500, 5219, 10460)
% DE 2 Wind power, Ampirion zonal load German’s Electricity Price 1 Hour (36500, 5219, 10460)

    \bottomrule
    % \bottomrule
    \end{tabular}
    }
    \end{small}
  \end{threeparttable}
  % \vspace{-5pt}
\end{table*}
For evaluation, following\cite{Timesnet}, we utilize the commonly used metrics in previous work\cite{Timesnet,liu2023itransformer}, including mean square error (MSE) and mean absolute error (MAE) for long-term forecasting. For short-term forecasting, following N-BEATS \citep{N-BEATS,Timesnet}, we adopt the symmetric mean absolute percentage error (SMAPE), mean absolute scaled error (MASE), and overall weighted average (OWA) as the metrics. The detailed calculations are as follows: 
\begin{align*} 
    \text{MAE} &= \frac{1}{H} \sum_{i=1}^H |\mathbf{X}_{i} - \widehat{\mathbf{X}}_{i}|, 
    &\text{MSE} &= \frac{1}{H} \sum_{i=1}^H (\mathbf{X}_{i} - \widehat{\mathbf{X}}_{i})^2, \\
    \text{SMAPE} &= \frac{200}{H} \sum_{i=1}^H \frac{|\mathbf{X}_{i} - \widehat{\mathbf{X}}_{i}|}{|\mathbf{X}_{i}| + |\widehat{\mathbf{X}}_{i}|},
    &
    \text{MAPE} &= \frac{100}{H} \sum_{i=1}^H \frac{|\mathbf{X}_{i} - \widehat{\mathbf{X}}_{i}|}{|\mathbf{X}_{i}|}, \\
    \text{MASE} &= \frac{1}{H} \sum_{i=1}^H \frac{|\mathbf{X}_{i} - \widehat{\mathbf{X}}_{i}|}{\frac{1}{H-m}\sum_{j=m+1}^{H}|\mathbf{X}_j - \mathbf{X}_{j-m}|},
    &
    \text{OWA} &= \frac{1}{2} \left[ \frac{\text{SMAPE}}{\text{SMAPE}_{\textrm{Naïve2}}}  + \frac{\text{MASE}}{\text{MASE}_{\textrm{Naïve2}}}  \right],
    \label{eq:metrics_detail}
\end{align*}
where $m$ is the periodicity of the data. $\mathbf{X},\widehat{\mathbf{X}}\in\mathbb{R}^{H\times C}$ are the ground truth and prediction results of the future with $H$ time pints and $C$ dimensions. $\mathbf{X}_{i}$ means the $i$-th future time point.
\section{Few-shot Learning}

The experiments in \cite{tsllm} indicate that LLM-based models do not benefit from pretrained large language models. To further evaluate the effectiveness of our multimodal model, we select three well-known end-to-end models as baselines. The results indicate that our method achieves comparable performance in 10\%-shot learning. However, due to the feature space gap between the language domain and time series data, our model fails in 5\%-shot learning.  Additionally, we observe that our model, when using a pretrained backbone, tends to overfit on the training and validation sets when trained with limited data, leading to lower performance on the test set.

% Table generated by Excel2LaTeX from sheet 'Sheet6'
\begin{table}[htbp]
  \centering
  \caption{Few-shot learning on Exchange dataset. }
    \begin{tabular}{cccccccccc}
    \toprule
    \multicolumn{2}{c}{Model} & \multicolumn{2}{c}{TimesCLIP} & \multicolumn{2}{c}{PatchTST} & \multicolumn{2}{c}{iTransformer} & \multicolumn{2}{c}{TimesNet} \\
    \multicolumn{1}{l}{Few-shot} & \multicolumn{1}{l}{Metric} & MSE$\downarrow$ & MAE$\downarrow$ & MSE$\downarrow$ & MAE$\downarrow$ & MSE$\downarrow$ & MAE$\downarrow$ & MSE$\downarrow$ & MAE$\downarrow$ \\
    \toprule
    \multirow{5}[1]{*}{20\%} & 96    & 0.121  & 0.241  & 0.087  & 0.208  & 0.100  & 0.226  & 0.153  & 0.275  \\
          & 192   & \underline{0.192}  & \underline{0.314}  & 0.174  & 0.300  & 0.196  & 0.319  & 0.260  & 0.366  \\
          & 336   & \underline{0.368}  & 0.447  & 0.290  & 0.393  & 0.370  & 0.446  & 0.417  & 0.472  \\
          & 720   & 0.938  & \underline{0.733}  & 0.717  & 0.647  & 0.926  & 0.734  & 0.962  & 0.747  \\
          & Avg.  & 0.405  & 0.434  & 0.317  & 0.387  & 0.398  & 0.431  & 0.448  & 0.465  \\
    \midrule
    \multirow{4}[2]{*}{10\%} & 96    & \underline{0.102}  & \underline{0.227}  & 0.089  & 0.210  & 0.114  & 0.241  & 0.152  & 0.275  \\
          & 192   & \textbf{0.187}  & \underline{0.316}  & 0.194  & 0.315  & 0.220  & 0.339  & 0.272  & 0.383  \\
          & 336   & \textbf{0.322}  & \textbf{0.417}  & 0.325  & 0.417  & 0.388  & 0.458  & 0.400  & 0.466  \\
          & Avg.  & \underline{0.204}  & \underline{0.320}  & 0.203   & 0.314  & 0.241  & 0.346  & 0.275  & 0.375  \\
    \midrule
    \multirow{3}[2]{*}{5\%} & 96    & 0.141  & 0.235  & 0.108  & 0.231  & 0.133  & 0.262  & 0.187  & 0.303  \\
          & 192   & 0.671  & 0.619  & 0.237  & 0.351  & 0.239  & 0.356  & 0.271  & 0.376  \\
          & Avg.  & 0.406  & 0.427  & 0.173  & 0.291  & 0.186  & 0.309  & 0.229  & 0.340  \\
    \bottomrule
    \end{tabular}%
  \label{tab:fewshot}%
\end{table}%

\section{Extra Ablation Study} \label{sec:app_ablation}
\subsection{Ablation of Visualization Preprocess} \label{sec:ablation_color_vis}
\begin{figure}[h]
% \vskip 0.2in
\begin{center}
\centerline{\includegraphics[width=0.8\textwidth]{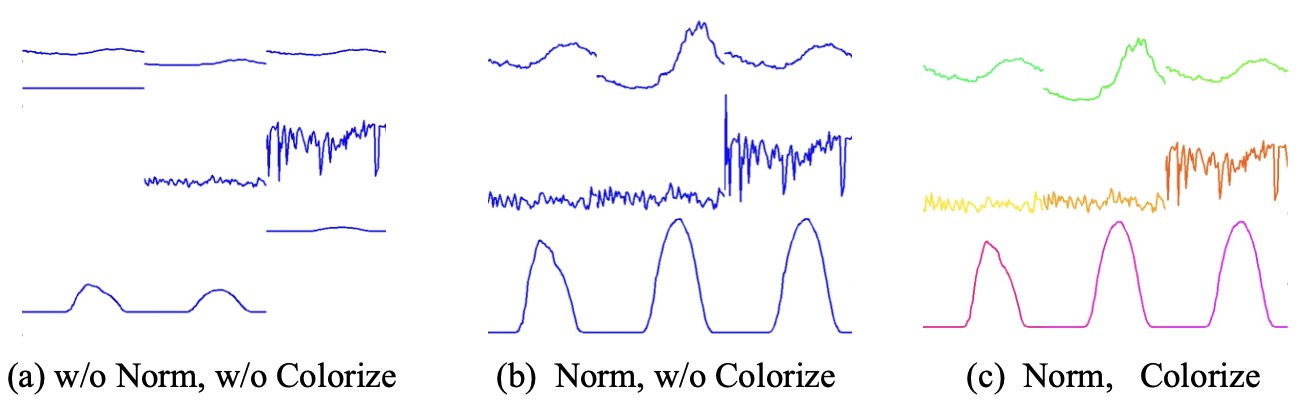}}
\vspace{-0.5em}
\caption{The visualization of Visualization Preprocess. Sample data is from Weather dataset.}
\vspace{-3em}
\label{fig:color}

\end{center}

\end{figure}

As shown in \cref{fig:color}, we visualize the results under two conditions: Norm and Colorize, as introduced in \cref{sec:vision-language}. \cref{fig:color}(a) indicates that the original numerical time series data spans a wide range. After normalization, as illustrated in \cref{fig:color}(b), we observe that variates with smaller values are magnified, making their periodicity more apparent. While the bottom three variates exhibit similar patterns, colorization helps the model distinguish different variates despite their structural similarity.

\subsection{Variate Fusion Strategy} \label{sec:ablation_fusion}
In this section, we conducted an ablation study on the fusion strategy for fusing the obtained $v_{CLS}$ to time series representation sequence as referred to in \cref{sec:generator}.

\noindent \textbf{Strategy explanation}. As introduced in \cref{sec:generator}, we propose a new variate fusion strategy to integrate the extracted correlative variate into the language representation of the time series. In detail, following  \cite{patchtst}, the patchify process includes a padding step, as detailed in \cref{eq:patchify}. The original length-\(T\) time series is padded to \(T + \text{Padding}\), where the padding length is half of the stride. Consequently, the last patch contains \(X^{T-\frac{S}{2}:T}\) and half padding data, while the second to last patch contains \(X^{T-S:T}\). Replacing the last token with \(v_{\text{CLS}}^{i}\) helps mitigate the influence of padding on the processing and avoids introducing additional parameters in the linear head. We conduct an ablation study of this specific fusion process in \cref{tab:variate}.

\noindent \textbf{Ablation of fusion strategy}.
We conducted ablation studies on three different fusion strategies designed to integrate the extracted correlative variate into the language representation of time series as the generator's input. These strategies include concatenating at the end of the feature sequence, replacing the $class$ token, and replacing the last feature of the language representation. As shown in \cref{tab:variate}, the results demonstrate that replacing the last feature of the language representation sequence is the most effective strategy for integrating extracted correlative variable features. As explained in \cref{sec:generator}, we think the padding step, introduced by \cite{patchtst}, employs "same" padding, and replicating the last length-\textit{Padding} of the time series features may compromise the model's forecasting accuracy.
% Table generated by Excel2LaTeX from sheet 'Sheet4'
\begin{table}[htbp]
  \centering
  \small
  \caption{Ablation study on variate fusion strategy }\label{tab:variate}%

  % \resizebox{0.9\columnwidth}{!}{
    \begin{tabular}{c|cc|ccc}
    \toprule

    \multirow{2}[2]{*}{Method} &\multicolumn{2}{c|}{Variate Fusion} &\multicolumn{3}{|c}{M4 Weighted Avg.} \\
    \cmidrule(lr){2-4} 
    \cmidrule(lr){4-6} 
    % \cmidrule(lr){5-5} 
    % \cmidrule(l){6-8} 
    &Strategy& Position & SMAPE$\downarrow$ &MASE$\downarrow$ & OWA$\downarrow$ \\
    
    \midrule
   \multirow{3}[2]{*}{Ours} &Replace  & First  &11.687 &1.563&0.840 \\

     & Concat   & End   &11.682 & 1.562 &0.839  \\
     \cmidrule(lr){2-6} 
     &Replace  & Last  &\textbf{11.642} &\textbf{1.560} &\textbf{0.837}  \\
% \cmidrule(lr){1-5}  
    \bottomrule
    \end{tabular}%
  % }
\end{table}%

% \begin{table}[htbp]
%   \centering
%   \caption{Ablation study of our proposed module.:{\color{red}experimenting}}\label{tab:ablation}%
%   \resizebox{\columnwidth}{!}{
%   % \setlength{\tabcolsep}{1.5pt}
%     \begin{tabular}{c|ccc|c|ccc}
%     \toprule  
%     \multirow{2}[2]{*}{Method} &\multicolumn{3}{|c|}{Vision Module} & {Variable}&\multicolumn{3}{|c}{M4 Weighted Avg.} \\
%     \cmidrule(lr){2-4} 
%     % \cmidrule(lr){5-5} 
%     \cmidrule(l){6-8} 
%     & $L_{cont}$& Col.  & Proj. & Selection & SMAPE$\downarrow$ &MASE$\downarrow$ & OWA$\downarrow$ \\
    
%     \midrule
%     \multirow{11}[2]{*}{Ours} & \graycross & \graycross & \graycross & \multirow{4}[2]{*}{\graycross} &  \\
%           & \ding{51} & \graycross & \graycross &  &  \\
%           & \ding{51} & \ding{51} & \graycross &   &  \\
%           & \ding{51} & \graycross  & \ding{51}&   &  \\
%           % & \graycross & \ding{51} & \ding{51} & \graycross &  \\
%           & \ding{51} & \ding{51} & \ding{51} &  &  \\
%     \cmidrule{2-8}
%           & \ding{51} & \graycross & \graycross & \multirow{4}[2]{*}{\ding{51}} &  \\
%           & \ding{51} & \ding{51} & \graycross &   &  \\
%           & \ding{51} & \graycross  & \ding{51}&   &  \\
%           % & \graycross & \ding{51} & \ding{51} & \graycross &  \\
%           & \ding{51} & \ding{51} & \ding{51} &   &  \\
%     \cmidrule{2-8}
%       & \ding{51} & \ding{51} & \ding{51} & \ding{51} & \\
%     \bottomrule
%     \end{tabular}%
% }
% \end{table}% 

\section{Experimental Details} \label{sec:detail_exp}
In this section, we provide more experimental details and analysis. For experiments result in \cref{tab:forecasting,tab:full_forecasting_results_m4}, if not otherwise stated, we adopt the baselines results from original paper and official code, except Time-LLM\cite{time-llm}. We noticed the short term forecasting results of other baselines in Time-LLM\cite{time-llm} are different with baselines' original paper, such as TimesNet have lower performance in \cite{time-llm}, although \cite{time-llm} claim their experimental setups followed \cite{Timesnet}. For fair comparison, we adopt Time-LLM results for long-term forecasting from \cite{tsllm}, which re-implements three LLM-based models and introduce a novel view that LLMs may not help time series analysis better. Additionally, we adopt the experiment results of Time-LLM for short-term forecasting from \cite{time-llm}, because these results have no advantage compared with other baselines' original results. Additionally, all the baselines that we reproduced are implemented based on the configurations of the original paper and official code. 
It is also notable that no multimodal-based methods are proposed for general time series analysis.  Please refer to our code, which shows more experimental details.

\subsection{short-term forecasting} \label{sec:short_detail}
% Table generated by Excel2LaTeX from sheet 'Sheet2'
 
We implement our model for short-term forecasting with the hyper-parameters in \cref{tab:m4_hyper}.
For M4 datasets, we conduct 17 uni-modality end-to-end baselines for short-term forecasting, which are classified into four architectures: (1)RNN-based models like LSTM \citeyearpar{hochreiter1997long} and S4 \citeyearpar{gu2021efficiently}; (2) CNN-based models, specifically TCN \citeyearpar{franceschi2019unsupervised} and TimesNet \cite{Timesnet}; (3) MLP-based models such as N-HiTS \citeyearpar{challu2023nhits}, N-BEATS \citeyearpar{N-BEATS}, LightTS \citeyearpar{campos2023lightts}, and DLinear \citeyearpar{DLinear}; (4) Transformer-based models including iTransformer \citeyearpar{liu2023itransformer}, Reformer \citeyearpar{kitaev2020reformer}, Informer \citeyearpar{zhou2021informer}, Pyraformer \citeyearpar{liu2022pyraformer}, Autoformer \citeyearpar{Autoformer}, FEDformer \citeyearpar{zhou2022fedformer}, Non-stationary Transformer \citeyearpar{liu2022non}, and \update{ETSformer \citeyearpar{woo2022etsformer}}. For PEMS, illness, and EPF datasets, we compared TimesCLIP with iTransformer and PatchTST. The full results are shown in \cref{tab:full_forecasting_results_m4,tab:pems,tab:epf}. 

% \input{tab/full_m4}
% Table generated by Excel2LaTeX from sheet 'small_short'

\begin{table}[htbp]
  \centering
  \caption{PEMS and illness, input 96 for PEMS, input 36 for illness}
  \small
    \begin{tabular}{c|c|cccccc}
    \toprule
    \multicolumn{2}{c|}{\multirow{2}[2]{*}{Model}} & \multicolumn{2}{c}{TimesCLIP} & \multicolumn{2}{c}{iTransformer} & \multicolumn{2}{c}{PatchTST} \\
        \multicolumn{2}{c|}{}   & \multicolumn{2}{c}{Ours} & \multicolumn{2}{c}{\citeyearpar{liu2023itransformer} } & \multicolumn{2}{c}{\citeyearpar{patchtst} } \\
    \cmidrule{1-8}
    \multicolumn{2}{c|}{Metric} & \multicolumn{1}{l}{MSE$\downarrow$} & \multicolumn{1}{l}{MAE$\downarrow$} & \multicolumn{1}{l}{MSE$\downarrow$} & \multicolumn{1}{l}{MAE$\downarrow$} & \multicolumn{1}{l}{MSE$\downarrow$} & \multicolumn{1}{l}{MAE$\downarrow$} \\
    \midrule
    \multirow{5}[2]{*}{\begin{sideways}PEMS08\end{sideways}} & 12    & 0.094  & 0.207  & 0.088  & 0.193  & 0.107  & 0.221  \\
          & 24    & \textbf{0.130 } & \textbf{0.244 } & 0.138  & 0.243  & 0.167  & 0.279  \\
          & 48    & \textbf{0.212 } & \textbf{0.276 } & 0.324  & 0.348  & 0.322  & 0.394  \\
          & 96    & \textbf{0.269 } & \textbf{0.319 } & 0.450  & 0.439  & 0.599  & 0.557  \\
          & Avg   & \textbf{0.176 } & \textbf{0.261 } & 0.250  & 0.306  & 0.299  & 0.363  \\
    \midrule
    \multirow{5}[2]{*}{\begin{sideways}PEMS04\end{sideways}} & 12    & \textbf{0.080 } & 0.201  & 0.081  & 0.189  & 0.109  & 0.223  \\
          & 24    & \textbf{0.088 } & \textbf{0.207 } & 0.100  & 0.212  & 0.179  & 0.293  \\
          & 48    & \textbf{0.100 } & \textbf{0.216 } & 0.135  & 0.248  & 0.344  & 0.414  \\
          & 96    & \textbf{0.124 } & \textbf{0.240 } & 0.167  & 0.279  & 0.615  & 0.580  \\
          & Avg   & \textbf{0.098 } & \textbf{0.216 } & 0.121  & 0.232  & 0.312  & 0.377  \\
    \midrule
    \multirow{5}[1]{*}{\begin{sideways}PEMS03\end{sideways}} & 12    & \textbf{0.069 } & 0.181  & 0.069  & 0.174  & 0.082  & 0.192  \\
          & 24    & \textbf{0.091 } & \textbf{0.206 } & 0.098  & 0.209  & 0.125  & 0.235  \\
          & 48    & \textbf{0.142 } & \textbf{0.258 } & 0.167  & 0.279  & 0.220  & 0.318  \\
          & 96    & \textbf{0.157 } & \textbf{0.257 } & 0.178  & 0.287  & 0.381  & 0.428  \\
          & Avg   & \textbf{0.115 } & \textbf{0.226 } & 0.128  & 0.237  & 0.202  & 0.293  \\
        \midrule
    \multirow{5}[1]{*}{\begin{sideways}PEMS07\end{sideways}} & 12    & \textbf{0.063 } & \textbf{0.175 } & 0.066  & 0.164  & 0.087  & 0.207  \\
          & 24    & \textbf{0.074 } & \textbf{0.183 } & 0.087  & 0.192  & 0.145  & 0.264  \\
          & 48    & \textbf{0.106 } & \textbf{0.204 } & 0.262  & 0.356  & 0.284  & 0.374  \\
          & 96    & \textbf{0.116 } & \textbf{0.210 } & 0.918  & 0.768  & 0.493  & 0.506  \\
          & Avg   & \textbf{0.358 } & \textbf{0.773 } & 1.334  & 1.480  & 1.009  & 1.351  \\
    \midrule
    \multirow{5}[2]{*}{\begin{sideways}illness\end{sideways}} & 24    & \textbf{2.007 } & \textbf{0.869 } & 2.317  & 0.941  & 2.199  & 0.889  \\
          & 36    & \textbf{2.160 } & \textbf{0.910 } & 2.199  & 0.950  & 2.362  & 0.917  \\
          & 48    & \textbf{1.849 } & \textbf{0.845 } & 2.243  & 0.964  & 2.028  & 0.876  \\
          & 60    & \textbf{1.927 } & \textbf{0.889 } & 2.287  & 0.990  & 1.960  & 0.888  \\
          & Avg   & \textbf{1.986 } & \textbf{0.878 } & 2.261  & 0.961  & 2.137  & 0.892  \\
    \bottomrule
    \end{tabular}%
  \label{tab:pems}%
\end{table}%

% % Table generated by Excel2LaTeX from sheet 'short-term'
% \begin{table}[htbp]
%   \centering
%   \caption{Add caption}
%     \begin{tabular}{cccccccc}
%     \toprule
%     \multicolumn{2}{c}{\multirow{2}[2]{*}{Model}} & \multicolumn{2}{c}{TimesCLIP} & \multicolumn{2}{c}{iTransformer} & \multicolumn{2}{c}{PatchTST} \\
%     \multicolumn{2}{c}{} & \multicolumn{2}{c}{Ours} & \multicolumn{2}{c}{} & \multicolumn{2}{c}{} \\
%     \midrule
%     \multicolumn{2}{c}{Metric} & \multicolumn{1}{l}{MSE\$\textbackslash{}downarrow\$ } & \multicolumn{1}{l}{MAE\$\textbackslash{}downarrow\$ } & \multicolumn{1}{l}{MSE\$\textbackslash{}downarrow\$ } & \multicolumn{1}{l}{MAE\$\textbackslash{}downarrow\$ } & \multicolumn{1}{l}{MSE\$\textbackslash{}downarrow\$ } & \multicolumn{1}{l}{MAE\$\textbackslash{}downarrow\$ } \\
%     \midrule

%   \label{tab:addlabel}%
% \end{table}%

\subsection{long-term forecasting} \label{sec:long_detail}
% To evaluate our method on Traffic and ECL, our CLIP-based model results in GPU memory consumption increasing with the number of variables. For instance, if we input 800 variates, the resulting language feature will have the shape $[B \times 800, 12, 512]$, requiring the processing of $B \times 800$ time series figures. This can lead to memory overflow. Hence, we adjust our model to embed numerical time series with the tokenizer, which is presented in \cref{sec:var}, and do not use a multimodal contrastive learning loss. Our model will only rely on the pretrained multimodal language model to long-term forecasting. As mentioned in \cref{sec:ablation_backbone}, mulimodal language model have the generazation ablitity to time series forecasting task. The experiment results are illustrated in \cref{tab:full_baseline_results}, our simplified model achieve the best performance on Traffic dataset. We think though efficent finetuning strategy, training our proposed model with constrastive loss will perform better. 
To evaluate our method on the Traffic and ECL datasets, our CLIP-based model results in GPU memory consumption increasing with the number of variables. For instance, if we input 800 variates, the resulting language feature will have the shape $[B \times 800, 12, 512]$, requiring the processing of $B \times 800$ time series figures, which can lead to memory overflow. 
To address this issue, we modify our model to embed numerical time series using the tokenizer, as described in \cref{sec:var}, and exclude the multimodal contrastive learning loss. The hyper-parameters are shown in \cref{tab:hyper_traffic}. As a result, the language feature shape becomes $[B, 800, 512]$, reducing the feature size by 12 times compared to the original design. Furthermore, our model relies solely on the pretrained multimodal language model for long-term forecasting. 

As discussed in \cref{sec:ablation_backbone}, multimodal language models exhibit strong generalization ability for time series forecasting tasks.
The experimental results, shown in \cref{tab:forecasting}, demonstrate that our simplified multimodal pretrained model still achieves the best performance on the Traffic dataset. We believe that with an efficient fine-tuning strategy, training our proposed model with contrastive loss would further improve its performance.


For Weather and Exchange dataset, we have different hyper-parameters setting depend on prediction length. The details are shown in \cref{tab:hyper_exchange,tab:hyper_Weather}.

\begin{table}[htbp]
  \centering
  
\begin{minipage}{0.45\textwidth}
  \centering
  \small
  \caption{\small Hyper-parameters on Traffic and ECL}
  \resizebox{\textwidth}{!}{
    \begin{tabular}{lcc}
    \toprule
    \multicolumn{1}{c}{\multirow{2}[2]{*}{Hyper-parameter}} & \multicolumn{2}{c}{Long-term Forecasting} \\
    \cmidrule{2-3}
          & Traffic &  ECL \\
    \midrule
    Optimizer & \multicolumn{2}{c}{AdamW\cite{adamw}} \\
    LR decay schedule & \multicolumn{2}{c}{Cosine Decaying to 0} \\
    Pred Length & \multicolumn{2}{c}{ 96, 192, 336, 720} \\
    train epoch & \multicolumn{2}{c}{50} \\
    Batch Size & \multicolumn{2}{c}{64} \\
    Pretrained Encoder LR & \multicolumn{2}{c}{1e-4} \\
    Others LR & \multicolumn{2}{c}{1e-4} \\
    early stop & \multicolumn{2}{c}{20} \\
    \bottomrule
    \end{tabular}%
}
  \label{tab:hyper_traffic}%
  
\end{minipage}
\hfill
\begin{minipage}{0.45\textwidth}
    \centering
    \small
    \caption{Hyper-parameters on Weather}
\resizebox{\textwidth}{!}{
    \begin{tabular}{l c c c c}
    \toprule
    \multicolumn{1}{c}{\multirow{1}[0]{*}{Hyper-parameter}} & \multicolumn{4}{c}{Weather (long-term)} \\
    % \cmidrule{2-3}
          % & Traffic &  ECL & & \\
    \midrule
    Optimizer & \multicolumn{4}{c}{AdamW\cite{adamw}} \\
    LR decay schedule & \multicolumn{4}{c}{Exponential Decay, $\gamma=0.5$} \\
    Pred Length &96 &192 &336 &720 \\
    train epoch & \multicolumn{4}{c}{10} \\
    Batch Size &8 &16 &64 &4 \\
    Pretrained Encoder LR &1e-6 &5e-6 &1e-6 &1e-6 \\
    Others LR & \multicolumn{4}{c}{1e-4} \\
    early stop & \multicolumn{4}{c}{3} \\
        $\lambda_1$ (weight of $L_{gen}$ ) & \multicolumn{4}{c}{1}\\
    $\lambda_2$ (weight of $L_{align}$) &  \multicolumn{4}{c}{0.1} \\
    \bottomrule

    \end{tabular}%
}
  \label{tab:hyper_Weather}%
  \end{minipage}

\end{table}

\begin{table}[htbp]
  \centering
  \begin{minipage}{0.45\textwidth}
    \centering
   \small
  \caption{Hyper-parameters on Exchange}
  \resizebox{\textwidth}{!}{
    % \begin{tabularx}{0.5\textwidth}{l c c c c}
    \begin{tabular}{l c c c c}
    \toprule
    \multicolumn{1}{c}{\multirow{1}[0]{*}{Hyper-parameter}} & \multicolumn{4}{c}{Exchange (long-term)} \\
    % \cmidrule{2-3}
          % & Traffic &  ECL & & \\
    \midrule
    Optimizer & \multicolumn{4}{c}{AdamW\cite{adamw}} \\
    LR decay schedule & \multicolumn{4}{c}{Exponential Decay, $\gamma=0.5$} \\
    Pred Length &96 &192 &336 &720 \\
    train epoch & \multicolumn{4}{c}{10} \\
    Batch Size &4 &8 &8 &4 \\
    Pretrained Encoder LR &1e-5 &1e-6 &5e-5 &5e-6 \\
    Others LR & \multicolumn{4}{c}{1e-4} \\
    early stop & \multicolumn{4}{c}{3} \\
    $\lambda_1$ (weight of $L_{gen}$ ) &\multicolumn{4}{c}{1}\\
    $\lambda_2$ (weight of $L_{align}$) & 0.5 & 0.1& 0.1& 0.9\\
    \bottomrule
    % \end{tabularx}%
    \end{tabular}%
    }
  \label{tab:hyper_exchange}%
  \end{minipage}
   \hfill
\begin{minipage}{0.45\textwidth}
  \centering
  \caption{\small Hyper-parameters on short forecasting.}
  \resizebox{\textwidth}{!}{
    \begin{tabular}{lc}
    \toprule
    Hyper-parameter & M4, PMES, illness, EPF \\
    \midrule
    Optimizer & AdamW\cite{adamw} \\
    LR decay schedule & Cosine   Decay to 0 \\
    train epoch & 100 \\
    Batch Size & 64 \\
    Pretrained Encoder LR & 1e-4 \\
    Others LR & 1e-3 \\
    $\lambda_1$ (weight of $L_{gen}$ ) & 1 \\
    $\lambda_2$ (weight of $L_{align}$) & 0.1 \\
    early stop & 30 \\
    \bottomrule
    \end{tabular}%
    }
  \label{tab:m4_hyper}%
  \end{minipage}
  % \hspac  e{0.2cm}

\end{table}

% \section{Few-shot Training}
\section{FLOPs and Parameters Analysis}

% We compare the efficiency of our model by FLOPs, Trainable parameters and total parameters with two transformer-based models, one CNN-based model, and two LLM-based models. Note that we implement the same model architecture, based on CLIP-ViT-B \cite{CLIP}, for each dataset, respectively, the induced little parameters is due to the more length learnable position embedding and class token. However, other baselines will be modified for different dataset, cause to the different number of model parameters. 
We compare the efficiency of our model in terms of FLOPs, trainable parameters, and total parameters against two Transformer-based models, one CNN-based model, and two LLM-based models. Note that we implement the same model architecture, based on CLIP-ViT-B \cite{CLIP}, across all datasets. The slight increase in parameters is due to the longer learnable position embeddings and the additional class token. In contrast, other baselines are modified for different datasets, leading to variations in model parameters.  
It is important to note that LLM-based models have a large number of total parameters; however, through Parameter-Efficient Fine-Tuning(PEFT)~\cite{han2024parameter}, they fine-tune only a small subset of parameters. This fine-tuning technique is also can be applied to our Transformer-based model to reduce the number of trainable parameters. However, since this work focuses on multimodal contrastive learning, we do not employ such techniques to accelerate training.

\begin{table}[h]
    \centering
    \caption{Comparison of FLOPs and parameters across different methods. Evaluate models with the prediction length set to 96. We adopt the results of Time-LLM and GPT4TS from \cite{tsllm}. Other baselines are evaluated based on their official codes and papers. The \textit{dim/Channel} denotes the hidden dimension of attention layer or CNN's channel depend on model's architecture. Due to Transformer-based model have different dimension of FFN, denoted as \textit{Dim. of FFN}, for instance, PatchTST has 2028 for Weather and 512 for Traffic, which cause different parameters.}
    \resizebox{\textwidth}{!}{
    \begin{tabular}{lc|ccc|ccc}
        \toprule
        Method & Dataset & \scalebox{0.9}{Enc. Layers}& \scalebox{0.9}{Dim/Channel} & \scalebox{0.9}{Dim. of FFN} & FLOPs(G)   & Trainable Param.(M)  & Total Param.(M)   \\
        \midrule
        \multirow{2}[1]{*}{TimesNet \citeyearpar{Timesnet}} &Exchange & 2 &64 &64& 4.54  & 4.71 & 4.71 \\
          &  Weather& 2  &32  &32 &1.13 & 1.19 & 1.19 \\
          &  Traffic& 2  &512 &512 &288.12  & 301.69 & 301.69 \\
        % PatchTST & 3.24  & 2.15 & 5.40 \\
        % iTransformer & 3.24  & 2.15 & 5.40 \\
        % Time-LLM & 4.18  & 3.02 & 6.87 \\
        \midrule
        \multirow{2}[1]{*}{PatchTST \citeyearpar{patchtst}} &Exchange & 2 &512 &2048 & 0.61  & 6.90  & 6.90  \\
         &Weather&2 & 512& 2048&  1.61  & 6.90& 6.90 \\
         &Traffic&2 & 512& 512&  33.52  & 3.76 & 3.76 \\
         \midrule
        \multirow{2}[1]{*}{iTransformer \citeyearpar{liu2023itransformer}} &Exchange & 2 &128 &128 & 2.77  & 0.22  & 0.22  \\
         &Weather &3 &512 &512 &  0.123  & 4.83 & 3.83 \\
         &Traffic &4 &512 &512 & 8.63  & 6.41 & 6.41 \\
         \midrule
         GPT4TS~\citeyearpar{gpt4ts}*  &Weather &\multicolumn{3}{c}{6 Layers of GPT2}   & - & -  & 86   \\
         Time-LLM~\citeyearpar{time-llm}* &Weather &\multicolumn{3}{c}{Llama-7B}   &- & -  & 6642   \\
        
         \midrule
        \multirow{2}[1]{*}{TimesCLIP (Ours)} & Exchange&  \multicolumn{3}{c|}{\multirow{3}{*}{CLIP-ViT-B (12 layers)}} &  39.44  & 65.85 & 153.31 \\
         & Weather &\multicolumn{3}{c|}{} & 103.21  & 65.87 & 153.32 \\
         & Traffic &\multicolumn{3}{c|}{} & 262.93  & 66.14 & 153.59 \\
        \bottomrule
    \end{tabular}
    }
    \label{tab:flops_params}
\end{table}

\section{Limitation and Further Improvement}
Our method is based on pretrained multimodal language encoder, a standard Transformer-based framework. However, the computational complexity of traditional Transformers, which scales as $O(n^2)$, poses a significant challenge when handling multiple variates as extremely long token sequences. This issue became apparent during our evaluation of the model on a traffic dataset, where we were compelled to omit the vision module to conserve GPU resources. Consequently, the number of trainable parameters and the demand for GPU resources increase as the token length grows with the number of variables. 
Fortunately, recent advancements have addressed these efficiency challenges; studies such as those on the Linear Transformer and lower parameter fine-tuning strategies like LoRA offer promising solutions. We think that designing a more efficient time series imaging strategy and alignment method to reduce resource requirements will extend this work as a foundational multimodal time series representation framework for more downstream tasks. 
  
Furthermore, mapping time‐series signals into a unified vision–language embedding space can benefit other modalities as well. For instance, repetitive action counting in videos can be framed as a time-series analysis problem \cite{hu2022transrac}. The proposed multimodal alignment framework can also be applied to other downstream applications such as data-centric AI~\cite{wang2025towards,ying2025survey,baiprivacy,gong2025agentic,gong2025sculpting,gong2025evolutionary,gong2025unsupervised,gong2025neuro,wang2024knockoff,wang2025llm,ying2023self,ying2024feature,ying2024unsupervised}, linguistics~\cite{ying2020sichuan,wang2022hierarchal}, business~\citep{rs1,wang2024llm,rs2,rs3,li2023sehf,wang2025enhanced}, and medicine~\cite{liu2019edta,wang2022successful,liu2024calorie,wang2024lcmdc,liu2024pth,li2024sade}.
Additionally, the reinforcement learning strategy~\cite{ying2025bridging}, reasoning process~\cite{wang2025efficient}, and the rule-based enhancement~\cite{bai2025brownian} could also be applied to enhance the robustness of the alignment process.

% \subsection{Architecture}

% \subsection{Parameter-Efficient Fine-Tuning}
% \subsection{Parameter Efficient}

\section{Related work} \label{sec:app_related}

\noindent\textbf{Time Series Forecasting with Language Models.}
Recent developments in multimodal large language models (LLMs), such as \cite{touvron2023llama,llava}, demonstrate the growing multimodal understanding capabilities of LLMs. For instance, \cite{gruver2024large} directly leverages the sequential processing capabilities of LLMs for numerical forecasting. Furthermore, several researchers have treated time series as a language \cite{jia2024gpt4mts, gpt4ts,time-llm,sun2023test,liu2024lstprompt}, employing pretrained large language models \cite{gpt3,touvron2023llama} to enhance forecasting methodologies. Notably, \cite{sun2023test} introduced TEST, which employs text prototype-aligned embeddings to augment the ability of LLMs to process time series data. Similarly, \cite{jin2023time} proposed Time-LLM, utilizing a novel training strategy named 'reprogramming,' designed to tailor LLMs' capabilities specifically for time series forecasting. However, a recent study \cite{tan2024language} suggests that LLM-based time series forecasting models might achieve better performance by replacing their LLM modules with a simpler encoder.

\noindent\textbf{Vision-Text Multimodal Learning}
In recent years, vision-text multimodal learning \cite{howto100m, sun2019learning, CLIP, videobert, actbert, LocalVTP, Xie2022Alignment, actionclip, wang2022long,coca,stockprice} has attracted increasing attention within the computer vision community. One of the most notable contributions is CLIP \cite{CLIP}, which successfully learns multimodal visual representations guided by natural language. CLIP utilizes a contrastive learning loss, specifically the InfoNCE loss \cite{InfoNCE}, to align visual representations with language representations within a shared multimodal space. The zero-shot ability of multimodal contrastive learning has spurred a multitude of follow-up works across various domains to align different modalities to a unified multimodal space, such as image domain \cite{li2023blip, coca,X-CLIP}, video domain \cite{videobert, actionclip, dong2023weakly, Xie2022Alignment,clip-event,sun2022long}, 3D domain \cite{stablediffusion,zhang2022pointclip,huo2025ct}, and audio domain \cite{guzhov2022audioclip,wu2022wav2clip}. Moreover, recent multimodal large language models\cite{llava,li2023videochat} continue to employ similar frameworks to integrate diverse modal representations with language representation spaces, unlocking the potential of LLMs. LLM routing can be applied to leverage the power of different LLMs \cite{wang2025mixllm}.

\noindent\textbf{Time Series Foundation Modals}.
Inspired by the concept of foundation models, such as LLMs, and moving away from training each model per task, some studies have also explored building foundational models for time series. For instance, \cite{goswami2402moment} introduced MOMENT, a family of pretrained time-series foundation models, and collected a new time series benchmark. \cite{liu2024unitime} introduced a unified model trained for cross-domain learning with natural language as domain instructions to provide domain-specific information. In contrast to \cite{patchtst}, which adapts pre-training strategies from neutral language processing and computer vision, \cite{liutimer,dong2024timesiam} introduced effective pre-training frameworks for time series forecasting. These efforts aim to construct robust time series foundation models.

%check list

% \newpage
% \input{sec/B_checklist}

% \section{Broader Impact}

% \textcolor{red}{new text} 

%%%%%%%%%%%%%%%%%%%%%%%%%%%%%%%%%%%%%%%%%%%%%%%%%%%%%%%%%%%%%%%%%%%%%%%%%%%%%%%
%%%%%%%%%%%%%%%%%%%%%%%%%%%%%%%%%%%%%%%%%%%%%%%%%%%%%%%%%%%%%%%%%%%%%%%%%%%%%%%

\end{document}